\documentclass{article}

\PassOptionsToPackage{numbers,sort&compress}{natbib}

\usepackage[main, final]{neurips_2026}

\usepackage[utf8]{inputenc} 
\usepackage[T1]{fontenc}    
\usepackage{hyperref}       
\usepackage{url}            
\usepackage{booktabs}       
\usepackage{bm}
\usepackage{amsfonts}       
\usepackage{nicefrac}       
\usepackage{microtype}      
\usepackage{xcolor}         
\usepackage{algorithm,algpseudocode} 
\usepackage{cancel}

\renewcommand{\v}[1]{\bm{#1}}       
\newcommand{\m}[1]{\bm{#1}}       
\usepackage{mbubble}
\mbubblestyle{ms}{color=teal,font={\tiny},label=ms,hlcolor=teal!15}

\title{Identifying Latent Neural Dynamics with Recognition-Parametrised Gaussian Process
Dynamical Systems}

\author{%
  Arielle Rosinski\thanks{These authors contributed equally.},\quad
  Qianqian Feng\footnotemark[1],\quad 
  Lior Fox,\quad
  Maneesh Sahani \\
  Gatsby Computational Neuroscience Unit\\
  University College London\\
  London, UK \\
  \texttt{\{arielle.rosinski.19, qianqian.feng, lior.fox\}@ucl.ac.uk}\\
  \texttt{maneesh@gatsby.ucl.ac.uk} \\
}

\usepackage{amsmath}

\usepackage{graphicx}
\setcitestyle{numbers,sort&compress}
\begin{document}

\maketitle

\begin{abstract}
Neural circuits are densely interconnected systems, and many neural computations are thought to be mediated by the dynamical evolution of activity within these recurrent networks, a hypothesis often referred to as ‘computation through dynamics’. Under this view, understanding computation within real neural populations depends on building models of the dynamical rules governing the temporal evolution of recorded high-dimensional neural activity. Most commonly, such models depend on explicit generative parameterisations of (i) the (nonlinear) intrinsic dynamical flow field and stochasticity, (ii) the (nonlinear) mapping from dynamical state to neuronal activity, and (iii) the variability of individual neural responses given that dynamical state. Misspecification of any of these components can bias estimates of dynamics and latent trajectories. In this paper, we circumvent generation-related issues by introducing the \textit{Recognition-Parametrised Gaussian Process} dynamics (RP-GPdyn) model. RP-GPdyn models the dynamical flow field using a nonparametric Gaussian Process-defined transition function and models the relationship of dynamics to neural activity implicitly using the Recognition-Parametrised Model (RPM) framework. We show that this generation-free approach uncovers meaningful, behaviourally relevant latent variables and dynamics from both synthetic and experimental datasets. 
\end{abstract}

\section{Introduction}\label{sec: introduction}
  Computation in brain circuits is shaped by a combination of input and recurrent processing, generating dynamical trajectories of population activity that influence downstream circuits, embody cognitive processes, and drive behaviour \citep{churchland2012neural, mante+al:2013, vyas2020computation}.
  Understanding such dynamical computation from neural recordings depends on tracking and visualising the collective circuit activity \citep{churchland+al:2007:conb, yu+al:2009:jneurophysiol}, identifying within it the computationally relevant distributed state \citep{saxena2019towards}, and determining the dynamical rules that govern its evolution \citep{macke2011empirical, duncker2021dynamics}.
  However, despite almost two decades of work, defining and estimating useful models of population dynamics remains an outstanding challenge in neuroscience for many reasons \citep{duncker2021dynamics}.
  First, theory suggests that the dynamical processes mediating computation may be essentially nonlinear.  Recurrent circuits can support multiple fixed points, continuous or approximate line and surface attractors, ring attractors, and limit cycles; providing mechanisms for memory, integration, categorical decisions, oscillations, and pattern generation \citep{hopfield:1982,amari:1977,seung:1996,zhang1996representation}.  At finer scales, dense nonlinear recurrence can induce chaotic or near-chaotic dynamics, and theoretical and machine-learning models have suggested that such regimes may be useful for flexible computation, transient amplification, sequence generation, and reservoir-like processing \citep{sompolinsky1988chaos,sussillo2009generating,laje2013robust}.
  Second, the internal dynamics may play out differently even when the same overt task is solved repeatedly \citep{churchland+al:2007:conb,  churchland2012neural,pandarinath2018inferring,vyas2020computation}.  Such variation may reflect different computational solutions, noise incorporated into the dynamics, or the superposition of unrelated processes carried within the same population \citep{stringer2019spontaneous}.
  Third, the recorded neurons may reflect the computational variables through a nonlinear mapping and with intrinsic noise that may not always be well-characterised by a Poisson process.

  Mitigating these profound challenges is the observation that, in many experimental settings, the behaviourally relevant dynamical variables appear to be far fewer than the number of neurons available in recordings \citep{churchland+al:2007:conb, cunningham2014dimensionality, gao2016linear}.
  This motivates analyses that model neural activity as the noisy observation of a low-dimensional latent dynamical system.
  The mapping from latent variables to population activity defines a differentiable dynamical ``manifold'' embedded within the space of neural firing patterns  \citep{duncker2021dynamics}.  Such low-dimensional descriptions have been useful in studies of motor control, decision making, navigation, working memory, and cognitive flexibility, where neural population activity often evolves along structured trajectories or geometries linked to task variables \citep{mante+al:2013,churchland2012neural,gallego2017neural,remington2018flexible,vyas2020computation}.
  
  Most latent dynamical models can be viewed as specifying three parametrised components: first, a low-dimensional dynamical flow, possibly stochastic and possibly driven by inputs; second, an embedding or emission function mapping latent states to neural activity; and third, an observation model describing how latent firing rates or intensities generate recorded data. Existing approaches differ in how they parametrise these components. Classical latent-variable models such as GPFA and linear state-space models provide tractable inference and interpretable low-dimensional trajectories \citep{yu+al:2009:jneurophysiol, cunningham2014dimensionality}. Extensions such as PLDS, nonlinear embeddings of linear dynamics, switching linear dynamical systems, recurrent switching state-space models, and neural-network-based sequential autoencoders increase flexibility while retaining varying degrees of interpretability \citep{macke2011empirical,gao2016linear,linderman2017bayesian,pandarinath2018inferring}.  More recent approaches, including DFINE, MARBLE, and Gaussian-process or switching-GP dynamical models, further emphasise nonlinear embeddings, nonlinear latent flows, cross-session or cross-animal alignment, and interpretable local dynamical structure \citep{abbaspourazad2024dfine,gosztolai2025marble,hu2024modeling}.

  Despite this progress, limitations remain.
  First, many models achieve expressive power at the cost of interpretability: flexible RNN or neural-ODE dynamics may predict neural activity well, but their vector fields, fixed points, local linearisations, and bifurcation structure often require substantial post-hoc analysis.
  Second, results depend on assumptions about the observation noise: spike-count distribution, calcium-imaging noise, or preprocessing pipeline. Misspecification of these components can distort latent trajectories, alter apparent dimensionality, or obscure the distinction between process noise and observation noise.
  Third, behaviour is treated as an external label to be decoded after the fact; whereas behavioural variables, task covariates, choices, movements, and internal states are often themselves coupled to the latent neural dynamics and should be incorporated directly into the model.
  Finally, modern experiments increasingly combine spikes, calcium signals, local field potentials, video-derived behavioural features, and task events, requiring models that can integrate heterogeneous data without forcing all observations into a single restrictive likelihood family.

  Here we introduce RP-GPdyn, a latent dynamical modelling framework designed to address these challenges.
  The model combines a nonparametric (Gaussian-process; GPdyn) dynamical prior with the recognition-parametrised model (RPM) framework.
  The RPM makes inference agnostic to the precise form of observation noise, learns complex nonlinear embeddings, and provides a natural route to combining heterogeneous data modalities, including neural recordings and behavioural covariates.
  The GPdyn component provides a flexible nonparametric model of nonlinear latent dynamics.
  To preserve interpretability of the latent dynamics, we condition the Gaussian-process flow field on learned fixed points and local Jacobians, following the strategy of \citet{duncker2019learning}. This yields a model that can express rich nonlinear stochastic dynamics while exposing dynamical features—fixed points, local stability, and flow geometry—that are central to mechanistic interpretation.
\section{Background}
\subsection{Gaussian Process-based Dynamical Systems}\label{sec: gp}
A GP defines a nonparametric distribution over functions in a manner that is amenable to Bayesian inference
\citep{williams2006gaussian}. 
In the GPdyn, this nonparametric framework is used to model the transition function of a nonlinear Markovian dynamical system.  
Let $\mathcal{Z} = \{\boldsymbol{z}_t \in \mathbb{R}^D\}_{t=1}^T$ represent a latent state-space trajectory, and $f_d(\cdot)$ for $d=1\dots D$ be draws from a GP prior.  Then the state transition model for the $d$th state coordinate is given by 
\begin{equation} \label{eq: f prior}
    z_t^{(d)} \mid \boldsymbol{z}_{t-1}, f_d \sim \mathcal{N}\big(f_d(\boldsymbol{z}_{t-1}), \sigma^2\big).
\end{equation}

GP inference is most straightforward when observations of the  inputs and (noisy) outputs of the target function are available; in the dynamical case, this would require state-space trajectories to be available for training \citep{ko2009gp, deisenroth2009analytic}.  
%
For a fully latent state process, as here, an alternative is to adopt the variational inducing-point framework of  \citet{titsias2009variational}; for dynamical applications see \citep{turner2010state, frigola2014variational, eleftheriadis2017identification, duncker2019learning, hu2024modeling}.
A variational distribution learnt over function values at inducing-point locations represents a posterior belief over dynamical transitions in a manner compatible with GP inference, and with fixed computational cost.  Introduce $M$ inducing-point locations $\m{\zeta} = \{  \v{\zeta}_i \in \mathbb{R}^{D} \}_{i=1}^M$ and associated outputs per dimension $\v{u}_d \in \mathbb{R}^{M}$.  An augmented GP prior over $f_d(\cdot)$ has the form
\begin{align}\label{eq: u prior}
 \v{u}_d &\sim \mathcal{N}(\boldsymbol{0}, K_{\zeta \zeta}) \\\nonumber
 f_d(\cdot) \mid \v{u}_d &\sim \mathcal{GP}\biggl(k(\cdot, \boldsymbol{\zeta}) K_{\zeta \zeta}^{-1}\m{u}_d, \; k(\cdot, \cdot) - k(\cdot, \m{\zeta}) K_{\zeta \zeta}^{-1}k( \m{\zeta}, \cdot) \biggr) 
\end{align}
where $k(\cdot,\cdot)$ is the covariance (kernel) function, and $K_{\zeta \zeta} := k(\m{\zeta}, \m{\zeta}) \in \mathbb{R}^{M \times M}$. The variational assumption then compresses the posterior belief over each $f_d$ into a finite Gaussian posterior on $\v{u}_d$ along with the \emph{prior} conditional $f_d | \v{u}_d$ of \eqref{eq: u prior}.  The locations $\m{\zeta}$ can be seen as variational parameters that can be optimised to improve the quality of this posterior approximation.

 \subsection{Interpretable Gaussian-Process Prior}\label{sec: interpretable GPs}
To facilitate direct access to fixed points of the dynamics $\mathcal{S} = \{\boldsymbol{s}_\ell \in \mathbb{R}^D\}_{\ell=1}^L$ and their local Jacobian matrices $\boldsymbol{J}^{(\ell)} \in \mathbb{R}^{D \times D} $, we define a GP prior conditioned explicitly on these parameters \citep[cf][]{bohner2018empirical, duncker2019learning}. In this formulation, the fixed point locations and Jacobians are hyperparameters $\boldsymbol{\theta} = \{ \boldsymbol{s}_\ell, \boldsymbol{J}^{(\ell)} \}$ that enforce the conditions $f(\boldsymbol{s}_{\ell}) = \boldsymbol{s}_{\ell} $ and $\left. \frac{\partial f_d(\boldsymbol{s})}{\partial s_k} \right|_{\boldsymbol{s} = \boldsymbol{s}_\ell} = \left[ \boldsymbol{J}^{(\ell)} \right]_{d,k}$. As a GP and its derivative
process are jointly also distributed as a GP, the sparse variational approximation framework in Section \ref{sec: gp} can be augmented by defining the joint covariance matrix between inducing points, fixed points and Jacobian matrices:
\begin{equation}
\mathsf{K}_{\zeta \zeta}^{\theta} = 
\begin{bmatrix}
K_{\zeta \zeta} & K_{\zeta \mathcal{S}} & \nabla_{2} K_{\zeta \mathcal{S}} \\
K_{\mathcal{S} \zeta} & K_{\mathcal{S}\mathcal{S}} &  \nabla_{2} K_{\mathcal{S}\mathcal{S}} \\
\nabla_{1} K_{\mathcal{S} \zeta} & \nabla_{1} K_{\mathcal{S}\mathcal{S}} & \nabla_{1}\nabla_{2} K_{\mathcal{S}\mathcal{S}}
\end{bmatrix}
=
\begin{bmatrix}
K_{\zeta \zeta} & \tilde{K}_{\zeta \mathcal{S}} \\
\tilde{K}_{\mathcal{S} \zeta} & \tilde{K}_{\mathcal{S}\mathcal{S}}
\end{bmatrix}
\end{equation}
where $K_{\mathcal{S}\mathcal{S}} := k(\mathcal{S}, \mathcal{S}) \in \mathbb{R}^{L \times L}$, and $\nabla_{i}$ indicates the derivative of the kernel with respect to its $i$-th input argument (see Appendix \ref{ap: rkhs block kernel} for a more exhaustive derivation of $\mathsf{K}_{\zeta \zeta}^{\theta}$). The prior in Equation \ref{eq: u prior} now becomes
\begin{align}\label{eq: augmented inducing point prior}
\v{u}_d \mid \boldsymbol{\theta} &\sim \mathcal{N}\biggl(\tilde{K}_{\zeta \mathcal{S}} \tilde{K}_{\mathcal{S} \mathcal{S}}^{-1}\boldsymbol{v}_d^{\theta}, \; K_{\zeta \zeta} - \tilde{K}_{\zeta \mathcal{S}} \tilde{K}_{\mathcal{S} \mathcal{S}}^{-1} \tilde{K}_{ \mathcal{S} \zeta} \biggr) \\ \nonumber
f_d(\cdot) \mid \v{u}_d, \boldsymbol{\theta} &\sim \mathcal{GP}\biggl(\boldsymbol{a}(\cdot)
\begin{bmatrix}
\boldsymbol{u}_d \\
\boldsymbol{v}^{\theta}_d
\end{bmatrix}, \; k(\cdot, \cdot) - \boldsymbol{a}(\cdot)\mathsf{K}_{\zeta \zeta}^{\theta} \boldsymbol{a}(\cdot)^{\top}  \biggr)
\end{align}
where 
$\boldsymbol{v}_d^{\theta} = \left[ {s}_{1}^{(d)}, \cdots, {s}_{L}^{(d)}, \boldsymbol{J}_{d,:}^{(1)}, \cdots, \boldsymbol{J}_{d,:}^{(L)} \right]^{\top}$ 
collects the fixed-points and derivatives corresponding to $f_d$, and $
\boldsymbol{a}(\cdot) = 
\left[ 
k(\cdot, \zeta) \quad 
k(\cdot, \mathcal{S})  \quad 
\nabla_{2} k(\cdot, \mathcal{S})
\right] 
\mathsf{K}_{\zeta \zeta}^{\theta^{-1}}$.

\subsection{Recognition-Parametrised Models}
Consider a set of related observed variables $\mathcal{X} = \{ \boldsymbol{x}_j \}_{j=1}^J$, assumed to be conditionally independent given latent variables
$\mathcal{Z} = \{ \boldsymbol{z}_k \}_{k=1}^K$. That is, the latents capture the statistical dependence amongst the observations and so their joint structure.  The full joint has the form
\begin{equation}
p(\mathcal{X}, \mathcal{Z}) = p(\mathcal{Z}) \prod_j p(\boldsymbol{x}_j \mid \mathcal{Z}) = p(\mathcal{Z}) \prod_j \bigg( p(\boldsymbol{x}_j)  \frac{p(\mathcal{Z} \mid \boldsymbol{x}_j) }{p(\mathcal{Z})} \bigg)   
\end{equation}
where the latter expression follows from Bayes' rule. The RPM \citep{walker2023unsupervised} proposes a normalised, semi-parametric hypothesis class to model such joint distributions.
Given a dataset $\mathbb{X}^{(N)}=\{ \mathcal{X}^{(n)}\}_{n=1}^N$ of $N$ data points, with $\mathcal{X}^{(n)} = \{ \boldsymbol{x}_j^{(n)} \}_{j=1}^J$, the RPM form is
\begin{equation}\label{eq: rpm default}
p_{\phi, \mathbb{X}^{(N)} }(\mathcal{X}, \mathcal{Z}) =    p(\mathcal{Z}) \prod_j \bigg( p_{0_j}(\boldsymbol{x}_j) \frac{g_{\phi_j}(\mathcal{Z} \mid \boldsymbol{x}_j) }{G_{\phi_j}(\mathcal{Z})} \bigg) 
\end{equation}
where 
$p_{0_j}(\boldsymbol{x}_j) = \frac{1}{N}\sum_n \delta(\boldsymbol{x}_j = \boldsymbol{x}_j^{(n)})$ is the empirical marginal on each observed variable; $p(\mathcal{Z})$ is a parametrised prior distribution on the latent
variables; the $g_{\phi_j}(\mathcal{Z} \mid \boldsymbol{x}_j)$ terms are normalised distributions, potentially parametrised as neural networks, referred to as the \textit{recognition factors}; and $G_{\phi_j}(\mathcal{Z}) = \int g_{\phi_j}(\mathcal{Z} \mid \boldsymbol{x}_j) p_{0_j}(\boldsymbol{x}_j) d\boldsymbol{x}_j = \frac{1}{N}\sum_n g_{\phi_j}(\mathcal{Z} \mid \boldsymbol{x}_j^{(n)})$, ensures proper normalisation of the RPM joint. 
The $\mathbb{X}^{(N)}$ in the subscript in \eqref{eq: rpm default} highlights the fact that the model parametrisation is itself data-dependent through $p_{0_j}$ and hence $G_{\phi_j}$. 
Thus, the RPM omits a parametric distribution over observations; instead learning focuses on recognition factors and prior.

State-space models offer a natural application of the RPM framework, as observations at different times are made conditionally independent by the latent state \cite{walker2023unsupervised,hromadka2026maximum}.  Following \citet{hromadka2026maximum}, we model time-series observations $\mathcal{X} = \{\boldsymbol{x}_t\}_{t=1}^T$ and underlying $D$-dimensional Markovian latents $\mathcal{Z} = \{\boldsymbol{z}_t\}_{t=1}^T$ with an RPM-defined joint given by 
\begin{equation} \label{eq: rpm ssm}
p_{\phi, \mathbb{X}^{(N)} }(\mathcal{X}, \mathcal{Z}) =    p(\boldsymbol{z}_1) \prod_{t=2}^T p(\boldsymbol{z}_t \mid \boldsymbol{z}_{t-1}) \prod_{t=1}^T
\bigg( p_{0}(\boldsymbol{x}_t) \frac{g_{\phi_t}(\boldsymbol{z}_t \mid \boldsymbol{x}_t) }{G_{\phi_t}(\boldsymbol{z}_t)} \bigg). 
\end{equation}
%
%

where the reference measure $p_0(x_t)$ can be constructed by pooling observations across both trials and time (time-pooled) $p_{0}(\boldsymbol{x}_t) = \frac{1}{NT}\sum_{nt} \delta(\boldsymbol{x}_t = \boldsymbol{x}_t^{(n)})$, or across trials separately at each time point (time-specific) $p_{0t}(\boldsymbol{x}_t) = \frac{1}{N}\sum_{n} \delta(\boldsymbol{x}_t = \boldsymbol{x}_t^{(n)})$. We defer discussion of the choice of reference measure to Section \ref{sec: maze} and Appendix \ref{ap: contrastive learning}.
In this formulation, the RPM framework offers a fully probabilistic, unsupervised approach, where the latent dynamics capture temporal dependence,
while the recognition factors link latent states to observations
without a parametric observation or noise model. Applied to neural data, there is no implicit Poisson assumption, and even noise processes that affect multiple neurons but do not have extended dynamical structure will be discounted appropriately.
RPM-based models have been shown to surpass comparable variational autoencoders (VAEs) for a wide range of tasks including image categorisation and video tracking of a moving agent \citep{walker2023unsupervised, hromadka2026maximum}.

\begin{figure}[h]
    \centering
    \includegraphics[width=\linewidth]{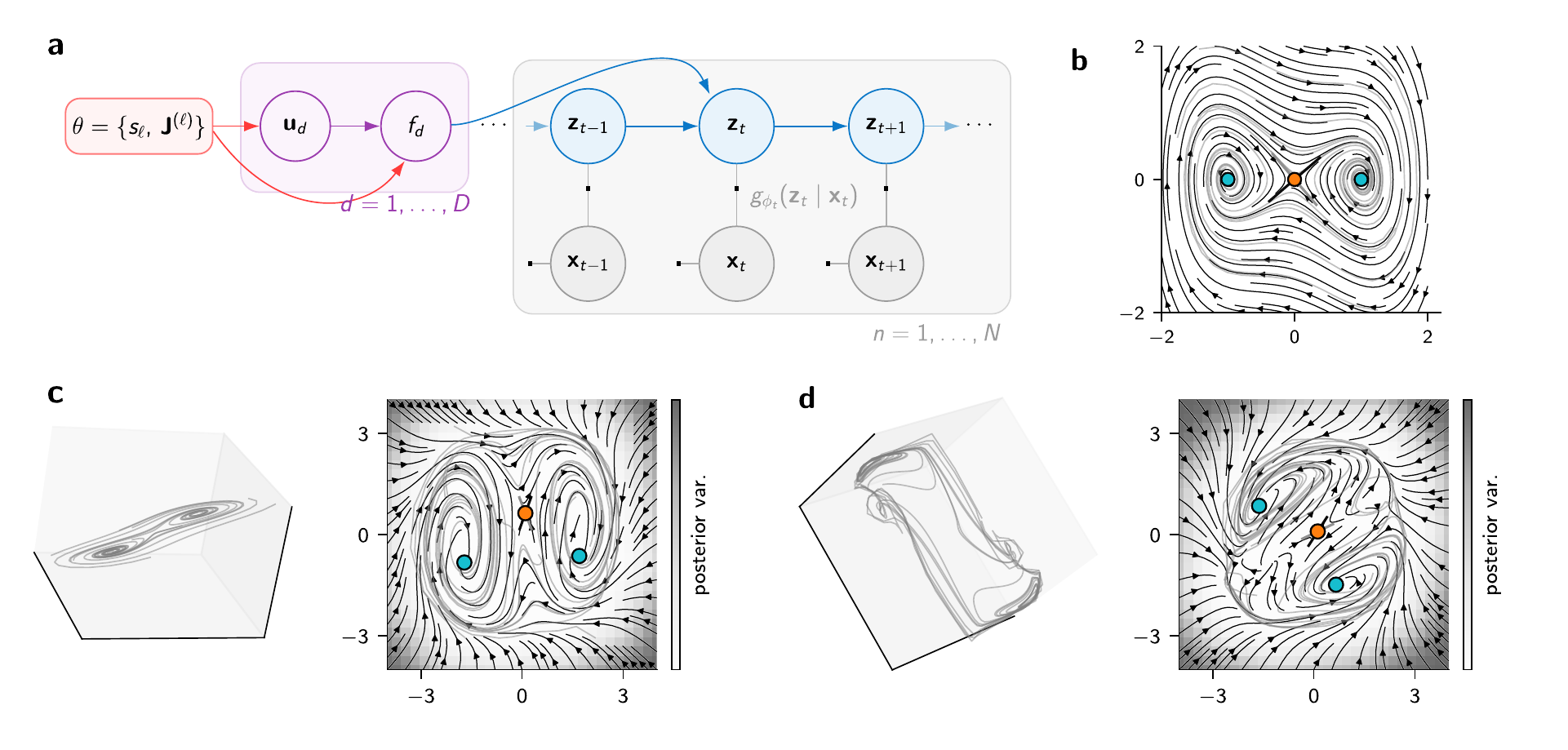}
    \caption{\textbf{(a)} RP-GPdyn defines a joint model over sequences of observations ($\boldsymbol{x}_t$), corresponding sequences of latent variables ($\boldsymbol{z}_t$), and global latent variables that define the flow field ($f_d$ and $\boldsymbol{u}_d$). The dynamics of the latent time series is modelled as a GP dynamics (Equation \eqref{eq: f prior}), where the GP is parametrised by a set of inducing points, as well as a set of fixed-points and their corresponding Jacobians, both of which are treated as (optimised) hyperparameters. The links between latents and observations are recognition-parametrised, with a parametric factor modelling $\boldsymbol{z}_t|\boldsymbol{x}_t$, and a non-parametric factor for the marginals of $\boldsymbol{x}_t$. %
    \textbf{(b)} Damped Duffing oscillator, an example of a nonlinear dynamical system in 2 dimensions. The dynamics are given by $\dot{z}_1=z_2$, $\dot{z}_2=z_1-z_1^3-0.3z_2$. The system is characterised by a saddle at the origin and two stable fixed points at $(\pm1,0)$. %
    \textbf{(c)} Trained on a linear embedding of the dynamics into three dimensions, RP-GPdyn successfully recovers the phase portrait of the system, as well as the fixed point stability structure. %
    \textbf{(d)} Same as \textbf{c}, but with the observations being embedded nonlinearly into 3 dimension, using a random neural network.
    %
    }
    \label{fig:model-overview}
\end{figure}

\section{The Recognition-Parametrised Gaussian Process
Dynamical Model}

Our \textit{Recognition-Parametrised Gaussian Process
dynamical model} (RP-GPdyn) is a latent time series model that combines a GP dynamical prior with recognition-parametrised links between observations and latents. The model structure is shown in Figure \ref{fig:model-overview}. A full description of the learning algorithm is given in Algorithm \ref{alg:rpgpdyn}.



\subsection{Variational Free Energy}\label{sec: free energy}
Inference and learning in RP-GPdyn use a Variational Bayes \citep{attias1999variational} approach, assuming a variational approximation to the posterior that factorises as  
\begin{equation}
q^{n}( \mathcal{Z}, f, \boldsymbol{u}) = q^{n}(\mathcal{Z}) \prod_d  p(f_d \mid \boldsymbol{u}_d, \boldsymbol{\theta}) q(\boldsymbol{u}_d).     
\end{equation}
where we define $q(\boldsymbol{u}_d) = \mathcal{N}(\boldsymbol{m}_d, \boldsymbol{\Gamma}_d)$ and $q^{n}(\boldsymbol{z}_t \mid \boldsymbol{z}_{t-1} ) = \mathcal{N}(\boldsymbol{A}_t^{(n)} \boldsymbol{z}_{t-1} + \boldsymbol{b}_t^{(n)}, \boldsymbol{Q}_t^{(n)})$, which provides a locally linear approximation to the nonlinear posterior over latents. The variational free energy follows from the log-joint over $\mathcal{X}$ and $\mathcal{Z}$ defined by the RPM model \eqref{eq: rpm ssm} combined with the inducing-point GP dynamics prior of \eqref{eq: f prior} and \eqref{eq: augmented inducing point prior}.  It can be written as
\begin{multline}\label{eq: free energy}
\mathcal{F}=\sum_{n=1}^N\left[
  \sum_{t=1}^T\left(
    \left\langle
      \log\!\frac{g_{\phi_t}(\boldsymbol{z}_t|\boldsymbol{x}_t^{(n)})}{G_{\phi_t}(\boldsymbol{z}_t)}
    \right\rangle_{q^n(\boldsymbol{z}_t)}
    \mskip -30mu+\log p_0(\boldsymbol{x}_t^{(n)})
  \right)
  - \left\langle\mathsf{KL}\!\left[q^n(\mathcal{Z})||p(\mathcal{Z}|f)\right]\right\rangle_{q(f)}
\right]
\\
-\sum_d\mathsf{KL}\!\left[q(\boldsymbol{u}_d)||p(\boldsymbol{u}_d|\boldsymbol{\theta})\right]
\end{multline}
%
where $q(f) = \prod_d  \int p(f_d \mid \boldsymbol{u}_d, \boldsymbol{\theta}) q(\boldsymbol{u}_d) d\boldsymbol{u}_d $, corresponding to the estimate of the dynamical transition, can be computed in closed form using properties of conditional Gaussian distributions. 
Note that $q(f)$ denotes the variational posterior over the GP transition function $f$, a global latent variable shared across all observations (as opposed to per-observation posteriors over latent variables $q(\boldsymbol{z}_n)$). 
As such, it reflects beliefs about the dynamics that are informed by the entire dataset, analogous to an empirical or learned prior. 
The derivation of the above expression for the free energy, as well as full details of learning and inference via variational Bayes expectation–maximisation (EM) are described in Appendix \ref{ap: lower bound}. Briefly, the variational distribution $q(\boldsymbol{u}_d)$, as well as updates for the Jacobian matrices $\boldsymbol{J}^{(\ell)}$ can be derived in closed form. The variational posterior on the latent variables $q^n(\boldsymbol{z}_t \mid \boldsymbol{z}_{t-1} )$ then corresponds to the output of a bi-directional recurrent neural network. This approach, which is similar to \citet{eleftheriadis2017identification}, enables the variational posterior at each time step to depend on both past and future observations, yielding a forward–backward smoothing approximation to the posterior over latent state sequences. 

In the form of Equation \ref{eq: free energy}, the free energy is intractable due to the expectation over the log Gaussian mixture term  $\langle \log G_{\phi_t}(\boldsymbol{z}_t)\rangle_{q^n(\boldsymbol{z}_t)}$. To address this issue, \citet{walker2023unsupervised} proposed several approximation techniques. The best empirical performance was achieved by the \textit{interior variational bound}, which lower-bounds the conventional free energy, and hence the log-likelihood. We thus select this approach here, and describe it in Appendix \ref{ap: interior bound}.

\section{Results}
\subsection{Synthetic Spiking Activity}
\begin{figure}s
    \centering
    \includegraphics[width=0.95\linewidth]{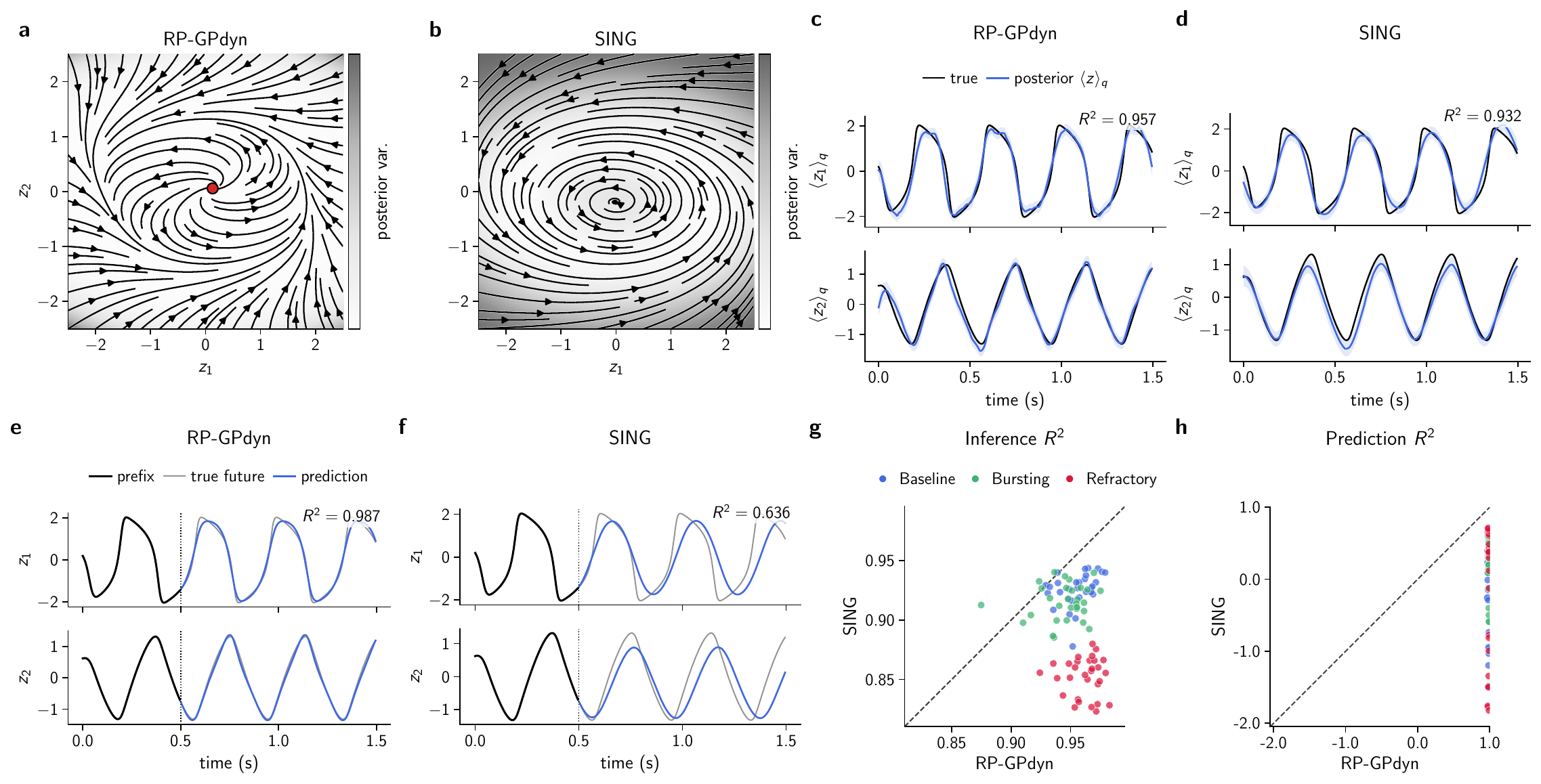}
    \caption{Van der Pol simulation. \textbf{(a)} RP-GPdyn flow field induced by the variational GP posterior
$q(f)$. Shading indicates the predictive variance of $q(f)$. \textbf{(b)} As \textbf{(a)}, for SING.  \textbf{(c)} For each latent dimension, example
RP-GPdyn posterior mean and $\pm$1 posterior standard deviation, shown against
the ground-truth latent trajectory. Because the latent coordinates are identifiable only up to a linear
transformation, the inferred posterior mean is aligned to the ground-truth latent space using a least-squares
linear map fit on the training latents. The $R^2$ for that trajectory is indicated. \textbf{(d)} As \textbf{(c)},  for SING. \textbf{(e)} RP-GPdyn prediction from a 0.5 s ground-truth prefix
(vertical dotted line), comparing the ground-truth future with
an autoregressive rollout under the learned transition posterior $q(f)$. The $R^2$ for that trajectory is indicated. \textbf{(f)} As \textbf{(e)}, for SING. \textbf{(g)} Latent inference $R^2$ comparisons for RP-GPdyn and SING for the baseline neural responses in \textbf{(a)}-\textbf{(f)} as well as bursting- and refractory-filtered activity. \textbf{(h)} As \textbf{(g)}, showing rollout predictions. }
    \label{fig: main vdp spikes}
\end{figure}

We applied RP-GPdyn to synthetic spike-train data generated from a latent Van der Pol oscillator ODE. Sampled latent states were mapped linearly through a random matrix to create inputs for 50 neurons. These inputs could then be combined with single-neuron history influences that either induced bursting or refractoriness, and were rectified with the softplus nonlinearity.
Overall, this noise process violates the widespread modelling assumption of memoryless Poisson emission noise.  Using 50 training trials and evaluated on 30 held-out trials, RP-GPdyn recovered the latent trajectories, dynamics, and phase portrait as shown in Figure \ref{fig: main vdp spikes}, including a fixed point consistent with the unstable spiral of the Van der Pol oscillator. 
Notably, comparisons with SING \cite{hu2026sing}, which also relies on GP-based transitions, though combined with explicit generative modelling, revealed greater robustness to history effects and overall superior performance for RP-GPdyn, with the largest gains in dynamics prediction (Figure \ref{fig: main vdp spikes}). Full details of data-generation and analyses are given in Appendix \ref{app:synthetic_history_effects}.



\subsection{Perceptual Decision Making}

Perceptual decision-making has been a classic case-study for investigating the dynamics of neural computation.
One well-studied mode of computation is through competition (via mutual inhibition) between distinct neural populations, resulting in a ``winner-take-all'' computation through bistability.

We tested the ability of the RP-GPdyn model to recover this bistable latent dynamical system from spiking observations.
Data was simulated following the model proposed by \citet{wang2002probabilistic} for the random dot motion task \cite{shadlen+newsome-2001:jphys}.
Concretely, the network comprises inhibitory and excitatory populations, with the latter
divided into 3 sub-populations: left-selective ($P=240$), right-selective ($P=240$), and non-selective neurons ($P=1120$).
The selective neurons receive external input that is modulated by the motion energy of the stimulus, as determined by the motion coherence of each trial.
We simulate activity from stimulus onset (after a 0.5s period of burn-in that was not included in the analysis).
The external stimulus is present for a period of 1 second, followed by another 0.5s post-stimulus period which is interpreted as the decision readout period.
Coherence levels were varied amongst 11 discrete settings between -51.2 and +51.2, including 0.

We fitted an RP-GPdyn model on the simulated spiking activity, with observations defined by the spike counts of each neuron in 10ms bins. The model was not given a priori knowledge about the existence of different subpopulations -- observations were full population vectors of all the excitatory neurons, selective and non-selective.
We take $\boldsymbol{z}$ to be two-dimensional, following the effective ``reduced'' (mean-field) description of this system \cite{wong+wang-2006:J-neuro}.

\begin{figure}[h]
  \centering

\includegraphics[]{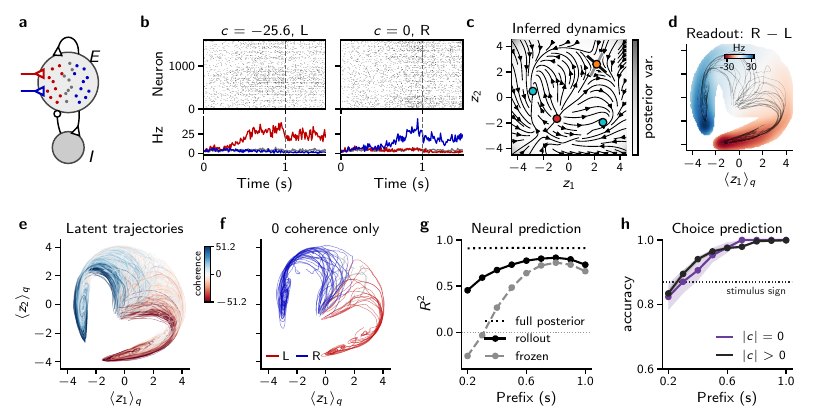}
\caption{Decision-making simulation. \textbf{(a)} Schematic structure of the spiking EI network. \textbf{(b)} 2 examples of trials from the dataset showing the spiking activity of the E population, and a ``ground truth'' readout of each of the left-, right-, and non-selective sub-populations average firing rate (red, blue, and grey respectively). 
\textbf{(c)} The latent dynamics learned by RP-GPdyn recover the bistability ``winner-takes-all'' dynamics implemented by the spiking network, with two stable fixed-points (blue) corresponding to decision states, separated by a separatrix that connects a saddle (orange) and an unstable fixed-point (red).
\textbf{(d)} Readout map for predicting left- and right-selective population firing-rates from the RP-GPdyn latent state.
\textbf{(e,f)} Inferred latent trajectories encode both the evidence accumulation indicated by their graded separation by trial coherence level, as well as the overall decision bistability, indicated by the ground truth trial decision, even on ambiguous zero-coherence trials. \textbf{(g)} Decoding selective population firing rates from inferred (posterior) and predicted (rollout) latent states. \textbf{(h)} Predicting trial choice from predicted latent state.
}

%
%
  \label{fig:decision-making}
\end{figure}

Trained on these data, RP-GPdyn recovered an accurate representation of the
bistable dynamics underlying the decision-making task (Figure
\ref{fig:decision-making}).
Inferred latent trajectories all start around a single point,
and converge to one of two ``decision areas'' in the latent space. 
While high-coherence trials converge rapidly to the decision area, low-coherence trials move further along a separatrix until driven, by fluctuations in the neural activity, to one of the two decision attractors.
Inferred latent states reliably predict the computational state of the circuit, that is, the instantaneous population firing-rates of the selective excitatory neurons.
Moreover, the learned dynamical system supports prediction of the evolution of these state variables from a partial prefix of observations.
Because the latent state reliably encodes these variables, it can also be used to predict the trial choice even from short prefixes of observations, including for ambiguous zero-coherence and low-coherence error trials.


\subsection{Persistent Activity in the Frontal Cortex}
\begin{figure}
    \centering
    \includegraphics[width=0.9\linewidth]{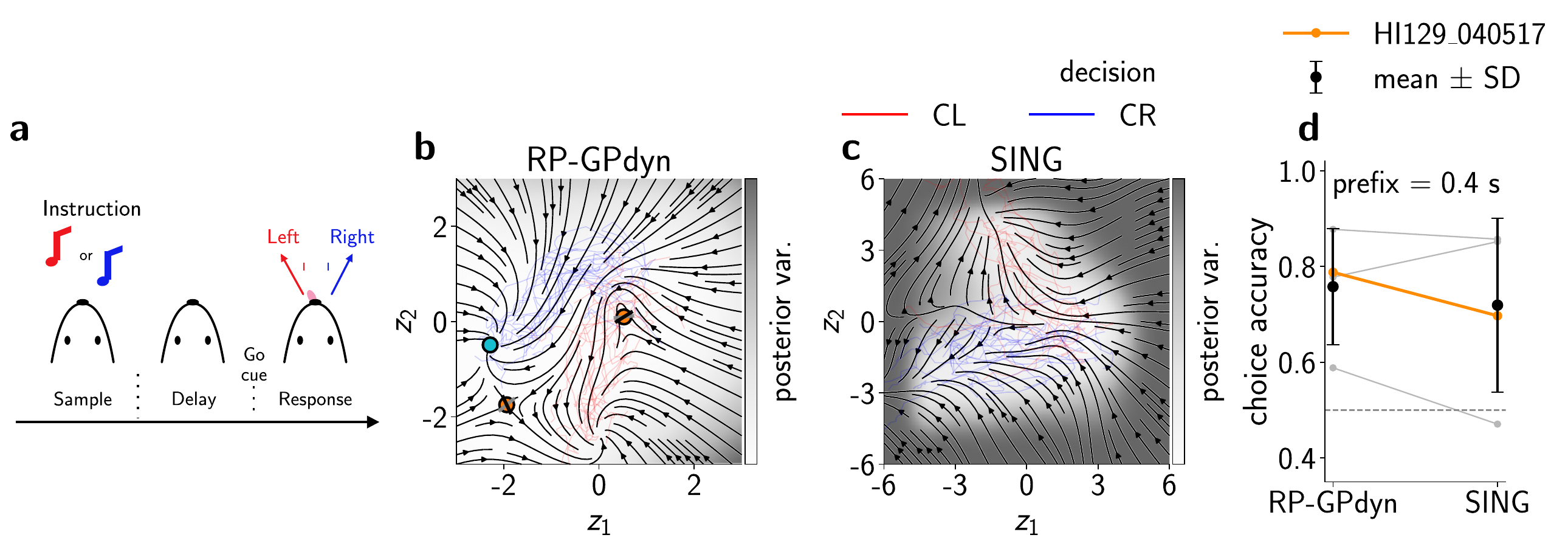}
    \caption{ALM data. (\textbf{a}) Auditory instruction tones are presented during the sample epoch, followed by a 2-s delay period. A brief auditory go cue then signals the response epoch, during which mice report the instructed choice by licking the left or right lick port. Figure adapted from \cite{inagaki2019discrete}. (\textbf{b}) RP-GPdyn-derived flow field overlaid with choice left (red) and right (blue)-corresponding latent trajectories learned from session \texttt{HI129 040517}. Fixed points are represented in blue (stable) and orange (saddle). We trained on 80\% of the trials and show results from the held-out 20\% test trials. (\textbf{c}) Same as (\textbf{b}) but for SING. (\textbf{d}) For RP-GPdyn and SING, we infer the latent trajectories up to a 0.4s prefix, then predict latent states forward using the learned dynamics. We then learn an instantaneous linear map to neural activity, based on which a logistic regression predicts choice. Different lines correspond to distinct recording sessions. The session pictured in \textbf{(b)} and \textbf{(c)} is displayed in orange.  }
    \label{fig: ALM paper}
\end{figure}

We next consider population activity in the anterior lateral motor cortex (ALM), a region critical to motor planning. We analysed extracellular activity in mouse ALM \cite{inagaki2018low, inagaki2019discrete} recorded while animals were presented with auditory instruction tones at one of two frequencies. Following a 2-second delay period, mice reported the instructed choice by licking either the left or right lick port (Figure \ref{fig: ALM paper}a). During the delay period, ALM population activity is approximately two-dimensional and exhibits persistent, choice-selective activity that predicts and instructs future actions \cite{inagaki2018low}. Notably, these responses have been shown to be consistent with discrete, bistable attractor dynamics, with each attractor corresponding to a distinct behavioural outcome \cite{inagaki2019discrete, finkelstein2021attractor}. 

Separate RP-GPdyn models were trained on single-trial delay-period neural responses from individual recording sessions spanning three animals (see details in Appendix \ref{ap: ALM}). Figure \ref{fig: ALM paper}b displays the learned flow field and latent trajectories for an example session, which support the previously proposed bistability mechanism. Additionally, predicted neural activity based on the learned dynamics accurately captured licking direction (Figure \ref{fig: ALM paper}d). By this classification-based metric, RP-GPdyn performed on par with SING (Figure \ref{fig: ALM paper}c, d), thereby supporting the reliability of our method. 

Note that, despite hinting at a bistable attractor structure, one fixed point appeared erroneous, consistently so across random seeds (not shown). The scarce nature of the data (59 neurons and 163 single trials) could imply that fixed point discovery is particularly challenging in this setting. Over-estimation of the fixed point number, followed by automatic relevance determination-based pruning as in \cite{duncker2019learning} could potentially enhance robustness in such cases. 

\begin{figure}
    \centering
    \includegraphics[width=\linewidth]{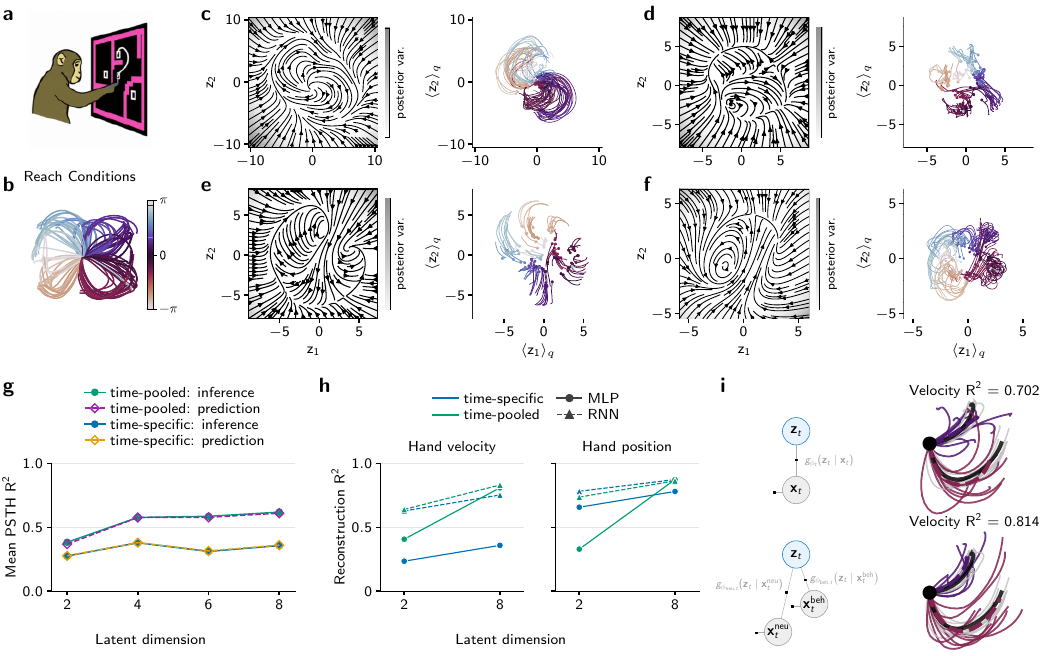}
    \caption{Motor cortex data.  \textbf{(a--b)} Schematic of the maze task \cite{pei2021neural} and reach conditions coloured by their mean reach directions.
    \textbf{(c--e)} RP-GPdyn models trained on single-trial neural activity with different reference measures and recognition models: \textbf{(c)} time-pooled measure, linear recognition; \textbf{(d)} time-specific measure, linear recognition; and \textbf{(e)} time-specific measure, MLP recognition. Each panel shows the learned posterior flow field (left) and latent trajectories inferred from condition-averaged neural activity (PSTHs; right).
    \textbf{(f)} Two-arm RP-GPdyn trained jointly on neural activity and hand velocity (graphical model shown at the bottom of panel~i), using a time-specific measure with linear recognition.
    \textbf{(g)} Mean neural PSTH $R^2$ on the official NLB test set, comparing true PSTHs with condition-averaged single-trial activity reconstructed from inferred posterior trajectories (inference) or GP transition rollouts (prediction).
    \textbf{(h)} Single-trial behaviour reconstruction $R^2$ decoded from the posterior means of neural-only models. We report the $R^2$ with both an instantaneous MLP decoder and a causal RNN decoder.
    \textbf{(i)} Left: Graphical model for processing neural-only versus multimodal (i.e., neural and behaviour) data. Right: example hand trajectories reconstructions (8D latent, linear recognition, MLP decoder). $R^2$ values are reported for the entire validation set.
    }
    \label{fig:mc_maze}
\end{figure}
\subsection{Behaviourally-Relevant Motor Cortex Neural Dynamics}
\label{sec: maze}
Finally, we applied RP-GPdyn to motor cortical population activity during
reaching, asking whether it could recover interpretable latent dynamics. 
This setting also allows us to examine how
the choice of reference measure shapes the recovered representation,
and whether incorporating behavioural observations during training
improves its behavioural relevance.

We analysed spiking activity from primary motor cortex (M1) and dorsal
premotor cortex (PMd) in the \texttt{MC\_Maze} dataset from the Neural
Latents Benchmark \citep{pei2021neural}. A monkey performed delayed
reaches through a virtual maze across 108 conditions, each specifying
a distinct reach geometry (Figure~\ref{fig:mc_maze}a,b). 
We focused on the movement epoch, during which previous work has identified coordinated, low-dimensional population dynamics
\citep{churchland2012neural,kao_single-trial_2015, sabatini_reach-dependent_2024}.

\noindent\textbf{Reference measures } RP-GPdyn can be flexibly configured to capture overall dynamical features (time-pooled, Figure \ref{fig:mc_maze}c) or condition-dependent dynamical features (time-specific, Figure \ref{fig:mc_maze}d,e) through the choice of reference measure $p_0(\boldsymbol{x}_t)$ (Equation \ref{eq: rpm ssm}). When the reference measure is time-pooled, the model is incentivised to capture all dynamical features through the latent $\boldsymbol{z}$. In contrast, a model with a time-specific reference measure captures dynamics that are specific to each individual trial or condition, since the overall temporal dynamics are already captured by the reference measure. 

These choices yielded qualitatively different two-dimensional
representations. RP-GPdyn trained on single-trial
neural activity with a time-pooled reference recovered a rotational
organisation reminiscent of the population dynamics identified by jPCA
\citep{churchland2012neural}
(Figure~\ref{fig:mc_maze}c). With a time-specific reference, latent
trajectories instead showed a more pronounced separation by reach
direction and a geometry more closely resembling the behavioural
trajectories (Figure~\ref{fig:mc_maze}d). This organisation was especially
apparent with a nonlinear recognition model
(Figure~\ref{fig:mc_maze}e). 

We asked whether the inferred latents support the prediction of neural activity beyond the neurons observed at test time. In the NLB dataset, test-time inference uses activity from 137 held-in neurons, while predictions are evaluated across these neurons and 45 held-out neurons whose activity is unavailable to the model at test time. During training, RP-GPdyn uses all neurons, but its amortised posterior receives only held-in activity. After decoding and condition averaging, posterior reconstructions achieved good neural PSTH $R^2$ scores on the official NLB test set (Figure~\ref{fig:mc_maze}g). We further tested whether the learned GP transition could reproduce the evolution of these latent states by generating autonomous rollouts initialised from inferred states. These rollouts achieved PSTH $R^2$ values close to those of full-posterior reconstructions, supporting the learned GP dynamics as a predictive description of latent evolution in real, noisy neural data.

To assess generalisation across movements, we trained on single-trial neural activity from 98 reach conditions
and evaluated on trials from the remaining 10 held-out conditions (Figure~\ref{fig:mc_maze}h). Latents learned from neural activity alone retained behaviourally relevant information: both instantaneous MLP and causal RNN readouts reconstructed hand velocity and position from the inferred trajectories (Figure~\ref{fig:mc_maze}h). Thus, behavioural structure emerged in the latent dynamics from neural data alone.

\noindent\textbf{Behaviour-guided training } When behavioural data or task-relevant hypotheses are available, RP-GPdyn can be  extended to incorporate these signals to guide the discovery of latent dynamics relevant to the variables of interest. The behaviour-guided model included a second recognition arm,
assuming conditional independence between neural and behavioural
observations given the latent state,
$\v x_t^{\mathrm{neu}} \perp \v x_t^{\mathrm{beh}} \mid \v z_t$
(Figure~\ref{fig:mc_maze}i).
During training, the latent state is encouraged to explain both modalities through the product of recognition factors,
$g_{\phi_{\mathrm{neu}, t}}(\v z_t \mid \v x_t^{\mathrm{neu}})
g_{\phi_{\mathrm{beh}, t}}(\v z_t \mid \v x_t^{\mathrm{beh}})$,
while the variational posterior $q_{\psi(\mathcal X^{\mathrm{neu}})}(\mathcal Z)$ remains fully amortised from neural activity alone. This allows behavioural signals to guide latent discovery without requiring behavioural observations at test time. Behaviour-guided training substantially improved the behavioural content of the inferred neural trajectories. When hand velocity was included as an additional training modality, posterior trajectories yielded more accurate single-trial reconstructions of hand kinematics on trials from held-out reach conditions (Figure~\ref{fig:mc_maze}f, i). Details of the experiments can be found in Appendix~\ref{ap:multi_arm_free_energy} and~\ref{ap: mcmaze}.


\section{Related work}
Circumventing the need for learning a generative model has previously been achieved by discriminative and contrastive approaches. Contrastive Predictive Coding (CPC; \cite{oord2018representation}) seeks to learn representations such that similar, or positive, examples (e.g., temporally related observations within the same sequence) are assigned similar representations, whereas dissimilar, or negative, examples (e.g., shuffled, randomly sampled observations) are assigned dissimilar representations. Notably, a relationship between RPM learning and contrastive approaches has been noted previously \cite{walker2023unsupervised} and is further discussed in Appendix \ref{ap: contrastive learning}. CEBRA, which extends and generalises the CPC-based learning objective, could be applied to the discovery of latent neural manifolds \citep{schneider2023learnable}. Yet, these methods do not learn a full probabilistic dynamics model, as they lack an explicit identification of the dynamical law 
of latent states, and do not provide posterior distributions over those states (but see \citep{eysenbach2024inference} for a probabilistic treatment of contrastive learning, albeit without explicit state-space dynamics).

\section{Discussion}
RP-GPdyn leverages probabilistic, nonlinear yet interpretable GP-based representations while avoiding explicit instantiation of the generative mapping from latent dynamics to neural activity. 
Moreover, the RPM structure lends itself naturally to multi-modal settings. Indeed, by treating neural data and behaviour as conditionally independent given the latents, the model is guided toward shared representations that jointly explain both modalities. While the present work has focused on synthetic spiking data and electrophysiological recordings, the generation-agnostic nature of RP-GPdyn makes it readily applicable to other neural data modalities. A particularly compelling direction for future work is calcium imaging, where the observation noise model is notoriously complex \citep{keeley2025improved}. 

The current model assumes autonomous dynamics, precluding the influence of exogenous feedforward inputs, such as contextual or sensory signals, which are known to shape neural dynamics alongside intrinsic recurrent connectivity. Still, in the settings studied here, this assumption appears reasonable. For instance, motor cortical dynamics during movement have been proposed to evolve near-autonomously \citep{pandarinath2018inferring, schimel2021ilqr, schimel2024and, churchland2012neural}. Inferring both latent dynamics and exogenous inputs simultaneously is challenging, as it requires teasing apart momentary variations attributable to the system's intrinsic transition function from those requiring external forcing inputs, an inherently degenerate problem. Recent attempts to address this include control-based inference using the iterative linear quadratic regulator \citep{schimel2021ilqr}, as well as resolving recurrent contributions through residual dynamics \citep{galgali2023residual}. Incorporating such approaches into RP-GPdyn is left for future work. 


Overall, within a single framework, RP-GPdyn addresses several longstanding neuroscience challenges such as nonlinear interpretable transitions, robustness to observation model misspecification, and principled incorporation of behavioural variables. Accordingly, it could constitute a powerful, flexible approach to studying neural computation across diverse recording modalities and experimental paradigms.

\section*{Acknowledgments}
This work was supported by the Gatsby Charitable Foundation (GAT3850, GAT4058) and the Simons Foundation (SCGB543039). We thank Lea Duncker, Samo Hromadka, Kai Biegun, and Sanvi Pal for
helpful discussions.

\bibliographystyle{unsrtnat}
\bibliography{ref}
\appendix

\section{Glossary of Notation}\label{ap: glossary}

\begin{table}[h]
\centering
\caption{Glossary of notation used throughout the paper.}
\label{tab: glossary}
\begin{tabular}{p{0.22\linewidth}p{0.72\linewidth}}
\toprule
\textbf{Symbol} & \textbf{Description} \\
\midrule
$N$ & Number of observed sequences or trials. \\
$T$ & Number of time steps per sequence. \\
$P$ & Observation dimension, e.g., number of recorded neurons or neural features. \\
$D$ & Latent state dimension. \\
$M$ & Number of inducing points used in the sparse GP approximation. \\
$L$ & Number of fixed points explicitly parameterizing the GP dynamics prior. \\
$\mathbb{X}^{(N)}$ & Dataset of observed sequences, $\mathbb{X}^{(N)}=\{\mathcal{X}^{(n)}\}_{n=1}^N$. \\
$\mathcal{X}^{(n)}$ & Observation sequence for trial $n$, $\mathcal{X}^{(n)}=\{\v{x}_t^{(n)}\}_{t=1}^T$. \\
$\v{x}_t^{(n)} \in \mathbb{R}^P$ & Observation at time $t$ in trial $n$. \\
$\mathcal{Z}^{(n)}$ & Latent state sequence for trial $n$, $\mathcal{Z}^{(n)}=\{\v{z}_t^{(n)}\}_{t=1}^T$. \\
$\v{z}_t^{(n)} \in \mathbb{R}^D$ & Latent state at time $t$ in trial $n$. \\
$f=(f_1,\ldots,f_D)$ & Latent transition function, with one GP output function per latent dimension. \\
$\sigma^2$ & Isotropic transition noise variance in the latent dynamics. \\
$k(\cdot,\cdot)$ & GP covariance kernel. \\
$\m{\zeta}=\{\v{\zeta}_m\}_{m=1}^M$ & Inducing point locations in latent space. \\
$\v{u}_d \in \mathbb{R}^M$ & Inducing targets for GP transition function $f_d$. \\
$\mathcal{S}=\{\v{s}_\ell\}_{\ell=1}^L$ & Fixed point locations of the latent dynamics. \\
$\m{J}^{(\ell)}$ & Jacobian matrix of the dynamics at fixed point $\v{s}_\ell$. \\
$\v{\theta}$ & Hyperparameters used to condition the interpretable GP prior, including fixed point constraints and Jacobians. \\
$\v{v}_d^\theta$ & Vector collecting fixed point and Jacobian constraints for GP output dimension $d$. \\
$p_0(\v{x}_t)$ & Empirical marginal distribution over observations at time $t$. \\
$g_{\phi_t}(\v{z}_t\mid \v{x}_t)$ & Recognition factor at time $t$.\\
$G_{\phi_t}(\v{z}_t)$ & Empirical recognition mixture.
\\
$q^n(\mathcal{Z})$ & Variational posterior over the latent trajectory for trial $n$. \\
$\m{A}_t^{(n)}$ & Locally linear transition matrix in the variational posterior. \\
$\v{b}_t^{(n)}$ & Locally linear offset in the variational posterior. \\
$\m{Q}_t^{(n)}$ & Variational transition covariance. \\
$q(\v{u}_d)$ & Variational posterior over inducing variables for GP output dimension $d$. \\
$\v{m}_d$ & Mean of $q(\v{u}_d)$. \\
$\m{\Gamma}_d$ & Covariance of $q(\v{u}_d)$. \\
$q(f)$ & Variational posterior predictive distribution over the latent dynamics. \\
\bottomrule
\end{tabular}
\end{table}

\section{Variational Lower Bound}\label{ap: lower bound}
\subsection{Lower Bound to the Marginal Log-Likelihood}
We derive the lower bound to the marginal log-likelihood in our model using Jensen’s
inequality
\begin{align}
\log p(\mathcal{X}^{(n)} \mid \v{\theta}) &=   \log \int q^n( \mathcal{Z}, f, \v{u}) \frac{p(\mathcal{X}^{(n)} \mid \mathcal{Z}) p(\mathcal{Z} \mid f) p(f \mid \boldsymbol{u}, \boldsymbol{\theta}) p(\boldsymbol{u} \mid \boldsymbol{\theta}) }{q^n( \mathcal{Z}, f, \v{u})} d\mathcal{Z} df d\boldsymbol{u}  \nonumber \\
&\geq   \int q^n( \mathcal{Z}, f, \v{u}) \log \frac{  p(\mathcal{X}^{(n)} \mid \mathcal{Z}) p(\mathcal{Z} \mid f) p(f \mid \boldsymbol{u}, \boldsymbol{\theta}) p(\boldsymbol{u} \mid \boldsymbol{\theta}) }{q^n( \mathcal{Z}, f, \boldsymbol{u})} d\mathcal{Z} df d\boldsymbol{u}  := \mathcal{F} 
\end{align}
where we have defined the following variational approximation to the posterior 
\begin{equation}
q^{n}( \mathcal{Z}, f, \boldsymbol{u}) = q^{n}(\mathcal{Z}) \prod_{d=1}^D  p(f_d \mid \boldsymbol{u}_d, \boldsymbol{\theta}) q(\boldsymbol{u}_d).  
\end{equation}

The bound, evaluated across all data points, can thus be re-expressed as
\begin{align}
\mathcal{F} 
&=\sum_{n=1}^N\left[\sum_{t=1}^T\left\langle \log p(\boldsymbol{x}_t^{(n)}\mid\boldsymbol{z}_t)\right\rangle_{q^n(\boldsymbol{z}_t)}-\left\langle\mathsf{KL}\!\left(q^n(\mathcal{Z})||p(\mathcal{Z}|f)\right)\right\rangle_{q(f)}\right]-\sum_d\mathsf{KL}\!\left(q(\boldsymbol{u}_d)||p(\boldsymbol{u}_d|\boldsymbol{\theta})\right)   \nonumber \\
&= \sum_{n=1}^N\left[\sum_{t=1}^T\left(\left\langle\log\!\frac{g_{\phi_t}(\boldsymbol{z}_t|\boldsymbol{x}_t^{(n)})}{G_{\phi_t}(\boldsymbol{z}_t)}\right\rangle_{q^n(\boldsymbol{z}_t)}+\log p_0(\boldsymbol{x}_t^{(n)})\right)-\left\langle\mathsf{KL}\!\left(q^n(\mathcal{Z})||p(\mathcal{Z}|f)\right)\right\rangle_{q(f)}\right]-\sum_d\mathsf{KL}\!\left(q(\boldsymbol{u}_d)||p(\boldsymbol{u}_d|\boldsymbol{\theta})\right).
\end{align}
where $\mathsf{KL}\!\left(q(\boldsymbol{u}_d)||p(\boldsymbol{u}_d|\boldsymbol{\theta})\right)$ corresponds to a Kullback–Leibler (KL) divergence between two Gaussian distributions, which evaluates to 
\begin{equation}
\mathsf{KL}\!\left(q(\boldsymbol{u}_d)||p(\boldsymbol{u}_d|\boldsymbol{\theta})\right) =\frac{1}{2} \left( \log \frac{\left| \boldsymbol{\Sigma}_{u} \right|}{\left| \boldsymbol{\Gamma}_d \right|} - M + \mathsf{Tr}[\boldsymbol{\Sigma}_{u}^{-1} \boldsymbol{\Gamma}_d]  + (\boldsymbol{\mu}_{u}^{(d)} - \v{m}_d)^{\top} \boldsymbol{\Sigma}_{u}^{-1} (\boldsymbol{\mu}_{u}^{(d)} - \v{m}_d) \right)    
\end{equation}
with $\boldsymbol{\mu}_{u}^{(d)} = \tilde{K}_{\zeta \mathcal{S}} \tilde{K}_{\mathcal{S} \mathcal{S}}^{-1}\boldsymbol{v}_d^{\theta}$ and $\boldsymbol{\Sigma}_{u} = \boldsymbol{\Sigma}_{u}^{(d)} = K_{\zeta \zeta} - \tilde{K}_{\zeta \mathcal{S}} \tilde{K}_{\mathcal{S} \mathcal{S}}^{-1} \tilde{K}_{ \mathcal{S} \zeta}$. The other terms in the free energy are further explained in Sections \ref{ap: interior bound} and \ref{ap: q(z) and KLqzpz}.

\subsection{Variational Interior Bound}\label{ap: interior bound}
In Section \ref{sec: free energy}, we have formulated the variational lower bound to the log likelihood 
\begin{align}
\mathcal{F} = \sum_{n=1}^N\left[\sum_{t=1}^T\left(\left\langle\log\!\frac{g_{\phi_t}(\boldsymbol{z}_t|\boldsymbol{x}_t^{(n)})}{G_{\phi_t}(\boldsymbol{z}_t)}\right\rangle_{q^n(\boldsymbol{z}_t)}+\log p_0(\boldsymbol{x}_t^{(n)})\right)-\left\langle\mathsf{KL}\!\left(q^n(\mathcal{Z})||p(\mathcal{Z}|f)\right)\right\rangle_{q(f)}\right]-\sum_d\mathsf{KL}\!\left(q(\boldsymbol{u}_d)||p(\boldsymbol{u}_d|\boldsymbol{\theta})\right).    
\end{align}
Yet, due to intractability of the log Gaussian mixture term  $\langle \log G_{\phi_t}(\boldsymbol{z}_t)\rangle_{q^n(\boldsymbol{z}_t)}$, we resort to the  \textit{interior variational bound} \citep{walker2023unsupervised, hromadka2026maximum}. To this end, we introduce auxiliary factors $\Tilde{g}_{tn}(\boldsymbol{z}_t)$ and compute the following via Jensen’s inequality
\begin{align}\label{eq: variational interior bound}
- \left\langle \log G_{\phi_t}(\boldsymbol{z}_t) \right\rangle_{q^{(n)}(\boldsymbol{z}_t)} 
&= - \left\langle \log \frac{G_{\phi_t}(\boldsymbol{z}_t) \Tilde{g}_{tn}(\boldsymbol{z}_t) q^{(n)}(\boldsymbol{z}_t)}{\Tilde{g}_{tn}(\boldsymbol{z}_t) q^{(n)}(\boldsymbol{z}_t)} \right\rangle_{q^{(n)}(\boldsymbol{z}_t)} \nonumber \\
&\geq \left\langle \log \frac{\Tilde{g}_{tn}(\boldsymbol{z}_t)}{q^{(n)}(\boldsymbol{z}_t)} \right\rangle_{q^{(n)}(\boldsymbol{z}_t)} - \log \underbrace{\int G_{\phi_t}(\boldsymbol{z}_t) \Tilde{g}_{tn}(\boldsymbol{z}_t) \, \mathrm{d}z_t}_{:=\Gamma_{\phi_t}^{tn}}.
\end{align}   

We therefore obtain the \textit{auxiliary free energy}
\begin{align}
\Tilde{\mathcal{F}}
&\overset{+c}{=} \sum_{n=1}^N \Bigg[
    \sum_{t=1}^T \Bigg(
        \left\langle
            \log \frac{
                g_{\phi_t}(\boldsymbol{z}_t \mid \boldsymbol{x}_t^{(n)})\,
                \Tilde{g}_{tn}(\boldsymbol{z}_t)
            }{
                q^{(n)}(\boldsymbol{z}_t)
            }
        \right\rangle_{q^{(n)}(\boldsymbol{z}_t)}
        - \log \Gamma_{\phi_t}^{tn}
    \Bigg)
    - \left\langle
        \mathsf{KL}\!\left(
            q^{(n)}(\mathcal{Z}) \,\|\, p(\mathcal{Z}\mid f)
        \right)
    \right\rangle_{q(f)}
\Bigg] \nonumber \\
&\quad
- \sum_d \mathsf{KL}\!\left(
    q(\boldsymbol{u}_d) \,\|\, p(\boldsymbol{u}_d \mid \boldsymbol{\theta})
\right)
\end{align}
In principle, the auxiliary factors $\Tilde{g}_{tn}(\boldsymbol{z}_t)$ can consist of parametrised (not necessarily normalised) functions, which can be optimised as part of the overall EM pipeline. Yet, previous work found an ansatz-based solution sufficient to achieve satisfactory results (see \citet{hromadka2026maximum, walker2023unsupervised} for discussion). Briefly, these authors note that the bound in Equation \ref{eq: variational interior bound} is tight when
$\Tilde{g}_{tn}(\boldsymbol{z}_t) \propto \frac{q^{(n)}(\boldsymbol{z}_t)}{G_{\phi_t}(\boldsymbol{z}_t)}$. Additionally, as discussed in \citet{walker2023unsupervised}
$G_{\phi_t}(\boldsymbol{z}_t) \to p_n(\boldsymbol{z}_t)
$
in the limit of large data under a correctly specified model. One can thus set 
$G_{\phi_t}(\boldsymbol{z}_t) \approx p(\boldsymbol{z}_t)$, such that $\Tilde{g}_{tn}(\boldsymbol{z}_t) = \frac{q^{n}(\boldsymbol{z}_t)}{p(\boldsymbol{z}_t)}$. Note that in the present case, deriving  $p(\boldsymbol{z}_t)$ from the GP-based joint over the full time-series is intractable. We therefore replace, for the definition of the auxiliary factors, $p(\boldsymbol{z}_t)$ by a standard normal distribution. This still results in a lower-bound, and we have found the modified auxiliary factors to perform well in practice. Moreover, since $p(\boldsymbol{z}_t \mid \boldsymbol{x}_t) \propto p(\boldsymbol{z}_t)\,p(\boldsymbol{x}_t \mid \boldsymbol{z}_t)$, we choose to define $g_{\phi_t}(\boldsymbol{z}_t \mid \boldsymbol{x}_t) = \frac{1}{Z^{tn}} p(\boldsymbol{z}_t)\, g_{\phi}^{\Delta}(\boldsymbol{z}_t \mid \boldsymbol{x}_t)$, where the $g_{\phi}^{\Delta}$ are parametrised (e.g., neural network-based), time-invariant, distributions. The final auxiliary free energy
can thus be re-expressed as 
\begin{align}\label{eq: final aux free energy}
\Tilde{\mathcal{F}}
&\overset{+c}{=} \sum_{n=1}^N \Bigg[
    \sum_{t=1}^T \Bigg(
        \left\langle
            \log \frac{
                g^{\Delta}_{\phi}(\boldsymbol{z}_t \mid \boldsymbol{x}_t^{(n)})\,
                \cancel{p(\boldsymbol{z}_t)} \cancel{q^{(n)}(\boldsymbol{z}_t)}
            }{
                \cancel{q^{(n)}(\boldsymbol{z}_t)} \cancel{p(\boldsymbol{z}_t)}
            }
        \right\rangle_{q^{(n)}(\boldsymbol{z}_t)}
        \underbrace{
            - \log \Gamma_{\phi_t}^{tn} - \log Z^{tn}
        }_{:= -\log \Tilde{\Gamma}^{tn}}
    \Bigg)
    \nonumber \\
&- \left\langle
        \mathsf{KL}\!\left(
            q^{(n)}(\mathcal{Z}) \,\|\, p(\mathcal{Z}\mid f)
        \right)
    \right\rangle_{q(f)}
\Bigg]
- \sum_d \mathsf{KL}\!\left(
    q(\boldsymbol{u}_d) \,\|\, p(\boldsymbol{u}_d \mid \boldsymbol{\theta})
\right).
\end{align}
To compute the $\log \Tilde{\Gamma}^{tn}$ term in Equation \ref{eq: final aux free energy}, we will rely on the following exponential family forms, which facilitate access to wide ranging probabilistic tools
\begin{align}
g(\boldsymbol{z}_t \mid \boldsymbol{x}_t^{(n)})
&= \exp\!\left( \eta(\boldsymbol{x}_t^{(n)})^\top t(\boldsymbol{z}_t) - \Phi\!\left(\eta(\boldsymbol{x}_t^{(n)})\right) \right), \\
q^{(n)}(\boldsymbol{z}_t)
&= \exp\!\left( (\eta_{q_t}^{(n)})^\top t(\boldsymbol{z}_t) - \Phi(\eta_{q_t}^{(n)}) \right), \\
p(\boldsymbol{z}_t)
&= \exp\!\left( \eta_{0t}^\top t(\boldsymbol{z}_t) - \Phi(\eta_{0t}) \right), \\
\Tilde{g}_{tn}(\boldsymbol{z}_t)
&= \exp\!\left( (\tilde{\eta}_t^{(n)})^\top t(\boldsymbol{z}_t) \right), \\
g^{\Delta}(\boldsymbol{z}_t \mid \boldsymbol{x}_t^{(n)})&= \exp\!\left( \eta^{\Delta}(\boldsymbol{x}_t^{(n)})^\top t(\boldsymbol{z}_t) - \Phi\!\left(\eta^{\Delta}(\boldsymbol{x}_t^{(n)})\right) \right).
\end{align}
where we have dropped the subscripts for brevity. Note that we can thus convey $g_{\phi_t}(\boldsymbol{z}_t \mid \boldsymbol{x}_t) \propto p(\boldsymbol{z}_t)\, g_{\phi}^{\Delta}(\boldsymbol{z}_t \mid \boldsymbol{x}_t)$ as $\eta(\boldsymbol{x}_t^{(n)}) = \eta_{0_t} + \eta^{\Delta}(\boldsymbol{x}_t^{(n)})$, and $\Tilde{g}_{tn}(\boldsymbol{z}_t) = \frac{q^{n}(\boldsymbol{z}_t)}{p(\boldsymbol{z}_t)}$ as $\Tilde{\eta}_t^{(n)} = \eta_{q_t}^{(n)} - \eta_{0_t} $. We then derive 
\begin{align}
\log \Gamma^{tn} &= \log \int d \boldsymbol{z}_t \Tilde{g}(\boldsymbol{z}_t) G(\boldsymbol{z}_t)   \nonumber \\
&= \log \frac{1}{N} \sum_{n'=1}^N \int d \boldsymbol{z}_t \Tilde{g}(\boldsymbol{z}_t)  g(\boldsymbol{z}_t \mid \boldsymbol{x}_t^{(n')}) \nonumber \\
&= \log \frac{1}{N} \sum_{n'=1}^N \int d \boldsymbol{z}_t \exp \left( (\tilde{\eta}_t^{(n)})^\top t(\boldsymbol{z}_t)   \right) \exp\!\left( \eta(\boldsymbol{x}_t^{(n')})^\top t(\boldsymbol{z}_t) - \Phi\!\left(\eta(\boldsymbol{x}_t^{(n')})\right) \right) \nonumber \\
&=  \log \frac{1}{N} \sum_{n'=1}^N \int d \boldsymbol{z}_t \exp \left( (\tilde{\eta}_t^{(n)}  + \eta(\boldsymbol{x}_t^{(n')}))^{\top}t(\boldsymbol{z}_t) - \Phi\!\left(\eta(\boldsymbol{x}_t^{(n')})\right)\right) \nonumber \\
&= \log \frac{1}{N} \sum_{n'=1}^N \exp \left( \Phi\left(\tilde{\eta}_t^{(n)}  + \eta(\boldsymbol{x}_t^{(n')})\right) - \Phi\!\left(\eta(\boldsymbol{x}_t^{(n')})\right)\right)
\end{align}
as well as
\begin{align}
\log Z^{tn} &= \log \int d \boldsymbol{z}_t p(\boldsymbol{z}_t) g^{\Delta} (\boldsymbol{z}_t \mid \boldsymbol{x}_t^{(n)}) \nonumber \\
&= \log \int d \boldsymbol{z}_t \exp\!\left( \eta_{0t}^\top t(\boldsymbol{z}_t) - \Phi(\eta_{0t}) \right)
\exp\!\left( \eta^{\Delta}(\boldsymbol{x}_t^{(n)})^\top t(\boldsymbol{z}_t) - \Phi\!\left(\eta^{\Delta}(\boldsymbol{x}_t^{(n)})\right) \right) \nonumber \\
&= \log \int d \boldsymbol{z}_t \exp \left( \left( \eta_{0t} + \eta^{\Delta}(\boldsymbol{x}_t^{(n)}) \right)^{\top} t(\boldsymbol{z}_t)-\Phi(\eta_{0t}) - \Phi\!\left(\eta^{\Delta}(\boldsymbol{x}_t^{(n)})\right)\right) \nonumber \\
&=  \Phi\left( \eta_{0t} + \eta^{\Delta}(\boldsymbol{x}_t^{(n)})\right)-\Phi(\eta_{0t}) - \Phi\!\left(\eta^{\Delta}(\boldsymbol{x}_t^{(n)})\right)
\end{align}
such that we can compute 
\begin{align}
\log \Tilde{\Gamma}^{tn}
&= \log \Gamma^{tn} + \log Z^{tn} \nonumber \\
&= 
    \Phi(\eta_{0t} + \eta^{\Delta}(\boldsymbol{x}_t^{(n)}))
    - \Phi(\eta_{0t})
    - \Phi(\eta^{\Delta}(\boldsymbol{x}_t^{(n)}))
 \nonumber \\
&\quad + \log \frac{1}{N} \sum_{n'=1}^N
\exp\!\big(
    \Phi(\tilde{\eta}_t^{(n)} + \eta(\boldsymbol{x}_t^{(n')}))
    - \Phi(\eta(\boldsymbol{x}_t^{(n')}))
\big) \nonumber \\
&= 
    \Phi(\eta_{0t} + \eta^{\Delta}(\boldsymbol{x}_t^{(n)}))
    - \Phi(\eta_{0t})
    - \Phi(\eta^{\Delta}(\boldsymbol{x}_t^{(n)}))
 \nonumber \\
&\quad + \log \frac{1}{N} \sum_{n'=1}^N
\exp\!\big(
    \Phi(\eta_{q_t}^{(n)} + \eta^{\Delta}(\boldsymbol{x}_t^{(n')}))
    - \Phi(\eta_{0t} + \eta^{\Delta}(\boldsymbol{x}_t^{(n')}))
\big).
\end{align}
\subsection{Inferring Latent Variables and Dynamics}\label{ap: q(z) and KLqzpz}
We define the variational distribution on latent variables as $q^{n}(\boldsymbol{z}_1) = \mathcal{N}(\boldsymbol{\mu}_0, \boldsymbol{Q}_0)$ and $q^{n}(\boldsymbol{z}_t \mid \boldsymbol{z}_{t-1} ) = \mathcal{N}(\boldsymbol{A}_t^{(n)} \boldsymbol{z}_{t-1} + \boldsymbol{b}_t^{(n)}, \boldsymbol{Q}_t^{(n)})$. The moments $\langle \boldsymbol{z}_t \rangle$, $\mathrm{Cov}[\boldsymbol{z}_t]$, and 
$\mathrm{Cov}[\boldsymbol{z}_t, \boldsymbol{z}_{t-1}]$ are then computed using standard properties of Gaussian distributions. The initial state parameters are learnable and randomly initialised. The parameters $\boldsymbol{A}_t^{(n)}$ and $\boldsymbol{b}_t^{(n)}$ are obtained via a bidirectional recurrent neural network with Gated Recurrent Units (GRUs) applied to the input sequences. Following \citet{eleftheriadis2017identification}, this parameterisation allows the variational posterior at each time step to condition on both preceding and succeeding observations, resulting in a forward--backward smoothing approximation. For simplicity, we assume $\boldsymbol{Q}_t^{(n)}$ is diagonal and shared across sequences. The KL divergence $\left\langle\mathsf{KL}\!\left(q^n(\mathcal{Z})||p(\mathcal{Z}|f)\right)\right\rangle_{q(f)}$ can then be computed as 
\begin{align}\label{eq: full klqzpz}
&\left\langle\mathsf{KL}\!\left(q^n(\mathcal{Z})||p(\mathcal{Z}|f)\right)\right\rangle_{q(f)} = \left\langle 
\log \frac{q^n(\boldsymbol{z}_1)}{p(\boldsymbol{z}_1)}
\right\rangle_{q^n(\boldsymbol{z}_1)}
+
\sum_{t=2}^{T}
\left\langle
\left\langle
\log
\frac{
q^n(\boldsymbol{z}_t \mid \boldsymbol{z}_{t-1})
}{
p(\boldsymbol{z}_t \mid \boldsymbol{z}_{t-1},  f)
}
\right\rangle_{q^n(\boldsymbol{z}_t,\boldsymbol{z}_{t-1})}
\right\rangle_{q(f)}
\nonumber\\[0.5em]
&\quad=
\frac{1}{2}
\left(
-\log |\boldsymbol{Q}_0|
- D
+ \mathrm{Tr}[\boldsymbol{Q}_0]
+ \boldsymbol{\mu}_0^{\top}\boldsymbol{\mu}_0
\right)
\nonumber\\
&\qquad\quad
+ \sum_{t=2}^{T}
\Bigg(
\frac{1}{2}
\log
\frac{|\sigma^2 \boldsymbol{I}|}{|\boldsymbol{Q}_t|}
-\frac{1}{2}
\Big(
\mathrm{Tr}\!\left[
\boldsymbol{Q}_t^{-1}
\left\langle
\boldsymbol{z}_t \boldsymbol{z}_t^{\top}
\right\rangle_{q(\boldsymbol{z}_t)}
\right]
\nonumber\\
&\qquad\qquad\qquad\qquad
-2\,\mathrm{Tr}\!\left[
\boldsymbol{Q}_t^{-1}\boldsymbol{A}_t
\left\langle
\boldsymbol{z}_{t-1}\boldsymbol{z}_t^{\top}
\right\rangle_{q(\boldsymbol{z}_t,\boldsymbol{z}_{t-1})}
\right]
+ \mathrm{Tr}\!\left[
\boldsymbol{A}_t^{\top}\boldsymbol{Q}_t^{-1}\boldsymbol{A}_t
\left\langle
\boldsymbol{z}_{t-1}\boldsymbol{z}_{t-1}^{\top}
\right\rangle_{q(\boldsymbol{z}_{t-1})}
\right]
\nonumber\\
&\qquad\qquad\qquad\qquad
-2\,\boldsymbol{b}_t^{\top}\boldsymbol{Q}_t^{-1}
\left(
\left\langle \boldsymbol{z}_t \right\rangle
-
\boldsymbol{A}_t \left\langle \boldsymbol{z}_{t-1} \right\rangle
\right)
+
\boldsymbol{b}_t^{\top}\boldsymbol{Q}_t^{-1}\boldsymbol{b}_t
\Big)
\nonumber\\
&\qquad\qquad
+\frac{1}{2\sigma^2}
\left\langle
\left\langle
\mathrm{Tr}\!\left[
\boldsymbol{z}_t \boldsymbol{z}_t^{\top}
\right]
+
\mathrm{Tr}\!\left[
f(\boldsymbol{z}_{t-1})f(\boldsymbol{z}_{t-1})^{\top}
\right]
-
2\,\boldsymbol{z}_t^{\top}f(\boldsymbol{z}_{t-1})
\right\rangle_{q(\boldsymbol{z}_t,\boldsymbol{z}_{t-1})}
\right\rangle_{q(f)}
\Bigg).
\end{align}
where $q(f) = \prod_d  \int p(f_d \mid \boldsymbol{u}_d, \boldsymbol{\theta}) q(\boldsymbol{u}_d) d\boldsymbol{u}_d = \mathcal{N}(\boldsymbol{\mu}_f, \boldsymbol{\Sigma}_f)$ with  
\begin{align*}
\boldsymbol{\mu}_f^{(d)} &= \boldsymbol{a}(\cdot) \begin{bmatrix}
\boldsymbol{m}_d \\
\boldsymbol{v}^{\theta}_d 
\end{bmatrix}  \\
\boldsymbol{\Sigma}_f^{(d)}&=k(\cdot, \cdot) - \boldsymbol{a}(\cdot)\mathsf{K}_{\zeta \zeta}^{\theta} \boldsymbol{a}(\cdot)^{\top} + \boldsymbol{a}(\cdot) \begin{bmatrix}
\boldsymbol{\Gamma}_d & \boldsymbol{0} \\
\boldsymbol{0} & \boldsymbol{0} \\
\end{bmatrix} \boldsymbol{a}(\cdot)^{\top}.
\end{align*} 

Note that evaluating the $\left\langle \left\langle \cdot \right\rangle_{q^{(n)}(\mathcal{Z})} \right\rangle_{q(f)}$ terms in Equation~\ref{eq: full klqzpz} requires computing expectations of kernel functions, their products, and their derivatives with respect to the variational distribution $q(\boldsymbol{z}_t) = \mathcal{N}(\boldsymbol{\mu}_t^{z}, \boldsymbol{\Sigma}_t^{z})$. For the RBF kernel $k(\boldsymbol{z}, \boldsymbol{z}')
= \sigma_f^2 \exp\!\left(
- \frac{1}{2\ell^2}
\|\boldsymbol{z} - \boldsymbol{z}'\|^2
\right)$, we obtain these as follows
\begin{align}
\left\langle \left\langle f_d(\boldsymbol{z}_{t-1}) \right\rangle_{q(\boldsymbol{z}_{t-1})} \right\rangle_{q(f)} &= \left\langle \boldsymbol{a}(\boldsymbol{z}_{t-1}) \right\rangle_{q(\boldsymbol{z}_{t-1})}\begin{bmatrix}
\boldsymbol{m}_d \\
\boldsymbol{v}^{\theta}_d 
\end{bmatrix} \\ \nonumber
&=  
\left\langle\left[ 
k(\boldsymbol{z}_{t-1}, \zeta) \quad 
k(\boldsymbol{z}_{t-1}, \mathcal{S})  \quad 
\nabla_{2} k(\boldsymbol{z}_{t-1}, \mathcal{S})
\right]  \right\rangle_{q(\boldsymbol{z}_{t-1})}
\mathsf{K}_{\zeta \zeta}^{\theta^{-1}}\begin{bmatrix}
\boldsymbol{m}_d \\
\boldsymbol{v}^{\theta}_d 
\end{bmatrix} 
\end{align}
where
\begin{itemize}
\item $\int q(\boldsymbol{z}_t)\, k(\boldsymbol{z}_t, \boldsymbol{\zeta}_m)\, d\boldsymbol{z}_t
= \sigma_f^2\left|\boldsymbol{I} + \tfrac{\boldsymbol{\Sigma}_t^{z}}{\ell^2}\right|^{-\frac{1}{2}}
\exp\!\left(
-\tfrac{1}{2}(\boldsymbol{\mu}_t^{z} - \boldsymbol{\zeta}_m)^\top
(\boldsymbol{\Sigma}_t^{z} + \ell^2\boldsymbol{I})^{-1}
(\boldsymbol{\mu}_t^{z} - \boldsymbol{\zeta}_m)
\right)
=: \Psi_1$

\item $\int q(\boldsymbol{z}_t)\, \nabla_{2} k(\boldsymbol{z}_t, \boldsymbol{s}_\ell)\, d\boldsymbol{z}_t
= \left[
(\boldsymbol{\Sigma}_t^{z} + \ell^2\boldsymbol{I})^{-1}
\!\left(\tfrac{\boldsymbol{\Sigma}_t^{z}}{\ell^2}\boldsymbol{s}_\ell + \boldsymbol{\mu}_t^{z}\right)
- \tfrac{\boldsymbol{s}_\ell}{\ell^2}
\right]\Psi_1$

\end{itemize}

\begin{align}
&\langle  \langle \mathsf{Tr}\left[  f_d(\boldsymbol{z}_{t-1}) f_d(\boldsymbol{z}_{t-1})^{\top} \right] \rangle_{q(f)}\rangle_{q(\boldsymbol{z}_{t-1})}  = \langle \boldsymbol{\Sigma}_f^{(d)} + \boldsymbol{\mu}_f^{(d)} \boldsymbol{\mu}_f^{(d)\top}\rangle_{q(\boldsymbol{z}_{t-1})}  \\ \nonumber
    &= \langle k(\boldsymbol{z}_{t-1}, \boldsymbol{z}_{t-1}) \rangle_{q(\boldsymbol{z}_{t-1})} + \mathsf{Tr} \left[ \left( \begin{bmatrix}
    \boldsymbol{\Gamma}_d & 0 \\
    0 & 0 
    \end{bmatrix} - \mathsf{K}_{\zeta \zeta}^{\theta} +\begin{bmatrix}
\boldsymbol{m}_d \\
\boldsymbol{v}^{\theta}_d 
\end{bmatrix} \begin{bmatrix}
\boldsymbol{m}_d \\
\boldsymbol{v}^{\theta}_d 
\end{bmatrix}^{\top} \right) \left\langle \alpha(\boldsymbol{z}_{t-1})^{\top}  \alpha(\boldsymbol{z}_{t-1})\right\rangle_{q(\boldsymbol{z}_{t-1})} \right]
\end{align}
where
\begin{itemize}
\item $\begin{aligned}[t]
\int &q(\boldsymbol{z}_t)\, k(\boldsymbol{z}_t, \boldsymbol{\zeta}_m)\,k(\boldsymbol{z}_t, \boldsymbol{\zeta}_{m'})\, d\boldsymbol{z}_t
= \sigma_f^4\left|\boldsymbol{I} + \tfrac{2\boldsymbol{\Sigma}_t^{z}}{\ell^2}\right|^{-\frac{1}{2}} \\
&\quad \times \exp\!\left(
-\tfrac{\|\boldsymbol{\zeta}_m-\boldsymbol{\zeta}_{m'}\|^2}{4\ell^2}
-\tfrac{1}{2}
\bigl(\boldsymbol{\mu}_t^{z} - \tfrac{\boldsymbol{\zeta}_m+\boldsymbol{\zeta}_{m'}}{2}\bigr)^\top
\bigl(\boldsymbol{\Sigma}_t^{z}+\tfrac{\ell^2}{2}\boldsymbol{I}\bigr)^{-1}
\bigl(\boldsymbol{\mu}_t^{z}-\tfrac{\boldsymbol{\zeta}_m+\boldsymbol{\zeta}_{m'}}{2}\bigr)
\right)
=: \Psi_2
\end{aligned}$

\item $\int q(\boldsymbol{z}_t)\,\nabla_{1}k(\boldsymbol{s}_\ell,\boldsymbol{z}_t)\,k(\boldsymbol{z}_t,\boldsymbol{\zeta}_m)\,d\boldsymbol{z}_t
= \Psi_2\!\left[
-\tfrac{\boldsymbol{s}_\ell}{\ell^2}
+\bigl(\boldsymbol{\Sigma}_t^{z}+\tfrac{\ell^2}{2}\boldsymbol{I}\bigr)^{-1}
\!\left(
\tfrac{\boldsymbol{\Sigma}_t^{z}(\boldsymbol{s}_\ell+\boldsymbol{\zeta}_m)}{2\ell^2}
+\tfrac{\boldsymbol{\mu}_t^{z}}{2}
\right)
\right]$

\item $\int q(\boldsymbol{z}_t)\,\nabla_{1}k(\boldsymbol{s}_\ell,\boldsymbol{z}_t)\,\nabla_{2}k(\boldsymbol{z}_t,\boldsymbol{s}_{\ell'})\,d\boldsymbol{z}_t
= \tfrac{\Psi_2}{\ell^4}
\Bigl(
\widetilde{\boldsymbol{\Sigma}}
+(\widetilde{\boldsymbol{\mu}}-\boldsymbol{s}_\ell)(\widetilde{\boldsymbol{\mu}}-\boldsymbol{s}_{\ell'})^\top
\Bigr)$, \\
where $\widetilde{\boldsymbol{\Sigma}}=\bigl((\boldsymbol{\Sigma}_t^{z})^{-1}+\tfrac{2}{\ell^2}\boldsymbol{I}\bigr)^{-1}$,
$\widetilde{\boldsymbol{\mu}}=\widetilde{\boldsymbol{\Sigma}}
\bigl(\tfrac{\boldsymbol{s}_\ell+\boldsymbol{s}_{\ell'}}{\ell^2}+(\boldsymbol{\Sigma}_t^{z})^{-1}\boldsymbol{\mu}_t^{z}\bigr)$
\end{itemize}

\begin{align}
\langle  
\langle 
\boldsymbol{z}_t^{\top} f(\boldsymbol{z}_{t-1}) 
\rangle_{q(f)} 
\rangle_{q(\boldsymbol{z}_t, \boldsymbol{z}_{t-1})}
&=
\mathsf{Tr} \Bigg[
\int  
q(\boldsymbol{z}_{t-1}) 
q(\boldsymbol{z}_t \mid \boldsymbol{z}_{t-1}) 
\boldsymbol{z}_t^{\top}    
\begin{bmatrix}
\boldsymbol{m}_1 & \cdots & \boldsymbol{m}_D \\
\boldsymbol{v}^{\theta}_1 & \cdots & \boldsymbol{v}^{\theta}_D
\end{bmatrix}^{\top}
\boldsymbol{a}(\boldsymbol{z}_{t-1})^{\top} 
\, d\boldsymbol{z}_t \, d\boldsymbol{z}_{t-1}
\Bigg]
\\[6pt]
&= \nonumber
\mathsf{Tr} \Bigg[
\begin{bmatrix}
\boldsymbol{m}_1 & \cdots & \boldsymbol{m}_D \\
\boldsymbol{v}^{\theta}_1 & \cdots & \boldsymbol{v}^{\theta}_D
\end{bmatrix}^{\top}
\Big\langle 
\boldsymbol{a}(\boldsymbol{z}_{t-1})^{\top} 
\boldsymbol{z}_{t-1}^{\top} 
\Big\rangle_{q(\boldsymbol{z}_{t-1})}
\boldsymbol{A}_t^{\top}
\Bigg]
\\[4pt]
&\quad
+
\mathsf{Tr} \Bigg[
\begin{bmatrix}
\boldsymbol{m}_1 & \cdots & \boldsymbol{m}_D  \nonumber \\
\boldsymbol{v}^{\theta}_1 & \cdots & \boldsymbol{v}^{\theta}_D
\end{bmatrix}^{\top}
\Big\langle 
\boldsymbol{a}(\boldsymbol{z}_{t-1})^{\top}
\Big\rangle_{q(\boldsymbol{z}_{t-1})}
\boldsymbol{b}_t^{\top}
\Bigg]
\end{align}
where
\begin{itemize}
\item $\int q(\boldsymbol{z}_t)\,\boldsymbol{z}_t\,k(\boldsymbol{z}_t,\boldsymbol{\zeta}_m)\,d\boldsymbol{z}_t
= \left((\boldsymbol{\Sigma}_t^{z})^{-1}\boldsymbol{\mu}_t^{z}+\tfrac{\boldsymbol{\zeta}_m}{\ell^2}\right)
\left((\boldsymbol{\Sigma}_t^{z})^{-1}+\tfrac{1}{\ell^2}\boldsymbol{I}\right)^{-1}
\Psi_1$

\item $\int q(\boldsymbol{z}_t)\,\boldsymbol{z}_t\,\nabla_{2}k(\boldsymbol{z}_t,\boldsymbol{s}_\ell)\,d\boldsymbol{z}_t
= \tfrac{\Psi_1}{\ell^2}
\Bigl(
\widetilde{\boldsymbol{\Sigma}}
+\widetilde{\boldsymbol{\mu}}\widetilde{\boldsymbol{\mu}}^\top
-\widetilde{\boldsymbol{\mu}}\boldsymbol{s}_\ell^\top
\Bigr)$, \\
where $\widetilde{\boldsymbol{\Sigma}}=\bigl((\boldsymbol{\Sigma}_t^{z})^{-1}+\tfrac{1}{\ell^2}\boldsymbol{I}\bigr)^{-1}$,
$\widetilde{\boldsymbol{\mu}}=\widetilde{\boldsymbol{\Sigma}}
\bigl((\boldsymbol{\Sigma}_t^{z})^{-1}\boldsymbol{\mu}_t^{z}+\tfrac{\boldsymbol{s}_\ell}{\ell^2}\bigr)$

\end{itemize}

\subsection{Closed Form Updates}\label{ap: closed form updates}
We can obtain closed-form updates for the inducing point means $\boldsymbol{m}_d$, inducing point covariances $\boldsymbol{\Gamma}_d$, and Jacobian matrices $\boldsymbol{J}_d$. 
The derivation follows \citet{duncker2019learning}, but is adapted to our discrete-time setting. 

For the covariance updates,
setting the derivative of the objective $\mathcal{F}$ with respect to $\boldsymbol{\Gamma}_d$ to zero yields
\begin{align}\label{eq:closed_form_gamma}
\boldsymbol{\Gamma}_d
=
\left(
\boldsymbol{\Sigma}_u^{-1}
+
\frac{1}{\sigma^2}
\sum_{n=1}^N \sum_{t=1}^T
\left[
\left\langle 
\boldsymbol{a}(\boldsymbol{z}_{t-1})^{\top} \boldsymbol{a}(\boldsymbol{z}_{t-1})
\right\rangle_{q^n(\boldsymbol{z}_{t-1})}
\right]_{:M,:M}
\right)^{-1}.
\end{align}
where $:M,:M$ indicates that we select the first $M \times M$ blocks. 

Similar to \citet{duncker2019learning}, we jointly parameterize the inducing means and Jacobian terms by forming the block matrix $
\boldsymbol{M}
=
\begin{bmatrix}
\boldsymbol{m}_1 & \cdots & \boldsymbol{m}_D \\
\boldsymbol{v}^{\theta}_1 & \cdots & \boldsymbol{v}^{\theta}_D
\end{bmatrix}$, where $[\boldsymbol{M}]_{:M, :M}$ and $[\boldsymbol{M}]_{M+L:,M+L:}$ index means and Jacobians, respectively. We can then re-write the quadratic terms in the KL divergence of the inducing points as  
\begin{align*}
\sum_d (\boldsymbol{\mu}_{u}^{(d)} - \boldsymbol{m}_d)^{\top}
\boldsymbol{\Sigma}_u^{-1}
(\boldsymbol{\mu}_{u}^{(d)} - \boldsymbol{m}_d)
=\begin{bmatrix}
\boldsymbol{\Sigma}_u^{-1} 
& -\boldsymbol{\Sigma}_u^{-1}\tilde{\boldsymbol{K}}_{\boldsymbol{\zeta}\mathcal{S}} \tilde{\boldsymbol{K}}_{\mathcal{SS}}^{-1} \\
\tilde{\boldsymbol{K}}_{\mathcal{SS}}^{-1} \tilde{\boldsymbol{K}}_{\mathcal{S}\boldsymbol{\zeta}}\boldsymbol{\Sigma}_u^{-1}
& \tilde{\boldsymbol{K}}_{\mathcal{SS}}^{-1} \tilde{\boldsymbol{K}}_{\mathcal{S}\boldsymbol{\zeta}}\boldsymbol{\Sigma}_u^{-1}
\tilde{\boldsymbol{K}}_{\boldsymbol{\zeta}\mathcal{S}} \tilde{\boldsymbol{K}}_{\mathcal{SS}}^{-1}
\end{bmatrix}=
\mathrm{Tr}\!\left[
\boldsymbol{M}^{\top} \widetilde{\boldsymbol{\Sigma}}\, \boldsymbol{M}
\right],
\end{align*}

Differentiating the $\boldsymbol{M}$-containing terms of the free energy then gives
\begin{align*}
\frac{\partial}{\partial \boldsymbol{M}}
\sum_d \mathsf{KL}\!\left(q(\boldsymbol{u}_d)\,\|\,p(\boldsymbol{u}_d \mid \boldsymbol{\theta})\right)
&=
\widetilde{\boldsymbol{\Sigma}}\, \boldsymbol{M},
\\
\frac{\partial}{\partial \boldsymbol{M}}
\sum_{n,t}
\left\langle 
\mathsf{KL}\!\left(q^n(\boldsymbol{z}) \,\|\, p(\boldsymbol{z} \mid f, \boldsymbol{a}) \right)
\right\rangle_{q(f)}
&=
\sum_{n,t}
\frac{1}{2\sigma^2}
\Big(
\left\langle \boldsymbol{a}(\boldsymbol{z}_{t-1})^{\top}\boldsymbol{a}(\boldsymbol{z}_{t-1}) \right\rangle \boldsymbol{M}
\\
&\quad 
- 2\, \left\langle \boldsymbol{a}(\boldsymbol{z}_{t-1})^{\top}\boldsymbol{z}_{t-1}^{\top} \right\rangle \boldsymbol{A}_t^{\top}
\Big).
\end{align*}

Setting the derivative of the full free energy to zero thus yields
\begin{align}\label{eq:closed_form_M}
\boldsymbol{M}
=
\left(
\widetilde{\boldsymbol{\Sigma}}
+
\frac{1}{2\sigma^2}
\sum_{n,t}
\left\langle \boldsymbol{a}(\boldsymbol{z}_{t-1})^{\top}\boldsymbol{a}(\boldsymbol{z}_{t-1}) \right\rangle
\right)^{-1}
\left(
\frac{1}{\sigma^2}
\sum_{n,t}
\left\langle \boldsymbol{a}(\boldsymbol{z}_{t-1})^{\top}\boldsymbol{z}_{t-1}^{\top} \right\rangle \boldsymbol{A}_t^{\top}
\right).
\end{align}

\subsection{Auxiliary Free Energy for Multi-arm Recognition}\label{ap:multi_arm_free_energy}
In the multi-arm variant of RP-GPdyn as described in Figure~\ref{fig:mc_maze}i, due to conditional independence, the free energy for such models with $J$ independent factors is simply
\begin{multline}
\mathcal{F}=\sum_{n=1}^N\Biggl[
  \sum_{t=1}^T\sum_{j=1}^J\left(
    \left\langle
      \log\!\frac{g_{\phi_{j,t}}(\boldsymbol{z}_t|\boldsymbol{x}_t^{(n)})}{G_{\phi_{j,t}}(\boldsymbol{z}_t)}
    \right\rangle_{q^n(\boldsymbol{z}_t)}
    \mskip-30mu+\log p_0(\boldsymbol{x}_t^{(n)})
  \right)
\\
  - \left\langle\mathsf{KL}\!\left[q^n(\mathcal{Z})\|p(\mathcal{Z}|f)\right]\right\rangle_{q(f)}
\Biggr]
-\sum_d\mathsf{KL}\!\left[q(\boldsymbol{u}_d)\|p(\boldsymbol{u}_d|\boldsymbol{\theta})\right]
\end{multline}

\begin{align}\label{eq: final aux free energy multiarm}
\widetilde{\mathcal F}_{J}
&\overset{+c}{=}
\sum_{n=1}^{N}
\Bigg[
\sum_{t=1}^{T}
\sum_{j=1}^{J}
\Bigg(
\left\langle
\log
\frac{
g^{\Delta}_{\phi_j}(\boldsymbol z_t \mid \boldsymbol x_{j,t}^{(n)})
\,\cancel{p(\boldsymbol z_t)}\,\cancel{q^{(n)}(\boldsymbol z_t)}
}{
\cancel{q^{(n)}(\boldsymbol z_t)}\,\cancel{p(\boldsymbol z_t)}
}
\right\rangle_{q^{(n)}(\boldsymbol z_t)}
-
\log \widetilde{\Gamma}_{j}^{tn}
\Bigg)
\nonumber \\
&\qquad
-
\left\langle
\mathsf{KL}\!\left(
q^{(n)}(\mathcal Z)\,\|\,p(\mathcal Z\mid f)
\right)
\right\rangle_{q(f)}
\Bigg]
-
\sum_{d=1}^{D}
\mathsf{KL}\!\left(
q(\boldsymbol u_d)\,\|\,p(\boldsymbol u_d\mid \boldsymbol\theta)
\right)
\\
&=
\sum_{n=1}^{N}
\Bigg[
\sum_{t=1}^{T}
\sum_{j=1}^{J}
\Bigg(
\left\langle
\log
g^{\Delta}_{\phi_j}(\boldsymbol z_t \mid \boldsymbol x_{j,t}^{(n)})
\right\rangle_{q^{(n)}(\boldsymbol z_t)}
-
\log \widetilde{\Gamma}_{j}^{tn}
\Bigg)
\nonumber \\
&\qquad
-
\left\langle
\mathsf{KL}\!\left(
q^{(n)}(\mathcal Z)\,\|\,p(\mathcal Z\mid f)
\right)
\right\rangle_{q(f)}
\Bigg]
-
\sum_{d=1}^{D}
\mathsf{KL}\!\left(
q(\boldsymbol u_d)\,\|\,p(\boldsymbol u_d\mid \boldsymbol\theta)
\right).
\end{align}
Here each arm contributes its own interior-bound correction
\[
\log \widetilde{\Gamma}_{j}^{tn}
=
\log \Gamma_{j}^{tn}
+
\log Z_{j}^{tn},
\]
where
\[
\Gamma_{j}^{tn}
=
\int
G_{\phi_j,t}(\boldsymbol z_t)\,
\widetilde g_{jtn}(\boldsymbol z_t)
\,d\boldsymbol z_t,
\qquad
Z_{j}^{tn}
=
\int
p(\boldsymbol z_t)\,
g_{\phi_j}^{\Delta}(\boldsymbol z_t \mid \boldsymbol x_{j,t}^{(n)})
\,d\boldsymbol z_t .
\]
And the new free energy results in a simple summation over the $J$ factors.

\section{Relation to Contrastive Approaches}\label{ap: contrastive learning}
A relationship between the RPM and contrastive approaches \cite{oord2018representation} that learn to contrast real observations from shuffled versions has been noted previously \cite{walker2023unsupervised}. It transfers to time-series, where the implicit joint over observations can be interpreted as weighting the empirical support 
\begin{equation}
P_{\phi, \mathbb{X}^{(N)}}(\mathcal{X}) = \prod_{t=1}^T p_{0t}(\boldsymbol{x}_t) \mathcal{W}_{\phi}(\mathcal{X}).
\end{equation}
Analogous to contrastive approaches, RPM learning thus maximises weights on plausible observations through the normaliser $\mathcal{W}_{\phi}(\mathcal{X})$
which, as the distribution is normalised, comes at
the expense of down-weighting implausible observations elsewhere. The precise meaning of shuffles here depends on the chosen base measure i.e., time-specific $p_{0t}(\boldsymbol{x}_t)=\frac{1}{N}\sum_n\delta\!\left(\boldsymbol{x}_t-\boldsymbol{x}_t^{(n)}\right)$ or time-pooled $p_{0}(\boldsymbol{x}_t)=\frac{1}{NT}\sum_n\sum_t\delta\!\left(\boldsymbol{x}_t-\boldsymbol{x}_t^{(n)}\right)$. For the time-specific base measure, the support is given by
\begin{align}
\prod_{t=1}^{T} p_{0t}(\boldsymbol{x}_t)
&=
\frac{1}{N^T}
\sum_{n_1,\ldots,n_T=1}^{N}
\prod_{t=1}^{T}
\delta\!\left(\boldsymbol{x}_t-\boldsymbol{x}_t^{(n_t)}\right).
\end{align}
such that genuine $\mathcal{X}$ consist of all observations belonging to the same trajectory, whereas for implicit shuffles different time positions can correspond to different trajectories. In contrast, the time-pooled support is 
\begin{equation}
\prod_{t=1}^{T} p_0(\boldsymbol{x}_t)
=
\frac{1}{(NT)^T}
\sum_{n_1,\ldots,n_T=1}^{N}
\sum_{s_1,\ldots,s_T=1}^{T}
\prod_{t=1}^{T}
\delta\!\left(
\boldsymbol{x}_t-\boldsymbol{x}_{s_t}^{(n_t)}
\right).
\end{equation}
such that now, genuine $\mathcal{X}$ consist of observation variables from one trial, in their original order, as opposed to shuffles, where each position can belong to a different trial and a different source time.

\section{Joint Kernel between Inducing points, Fixed points and Jacobian Matrices}\label{ap: rkhs block kernel}

We derive the joint covariance $\mathsf{K}_{\zeta\zeta}^{\theta}$ from Section~\ref{sec: interpretable GPs} \citep{duncker2019learning, bohner2018empirical}. Let $k$ be a positive-definite kernel with reproducing kernel Hilbert space (RKHS) $\mathcal{H}_k$. The reproducing property gives $h(\boldsymbol{z}) = \langle h, k(\cdot,\boldsymbol{z})\rangle_{\mathcal{H}_k}$ for all $h\in\mathcal{H}_k$. Differentiating both sides with respect to $z_i$, and using continuity of the inner product, yields the \emph{derivative reproducing property}:
\begin{equation}
    \frac{\partial h(\boldsymbol{z})}{\partial z_i} = \left\langle h,\, \frac{\partial k(\cdot,\boldsymbol{z})}{\partial z_i}\right\rangle_{\mathcal{H}_k}.
\end{equation}
%

In our case, $\partial k(\cdot,\boldsymbol{z})/\partial z_i \in \mathcal{H}_k$. The cross-covariances between function values and partial derivatives are given by
\begin{align}
    \operatorname{Cov}\!\left(f_d(\boldsymbol{z}),\, \frac{\partial f_d(\boldsymbol{z}')}{\partial z'_j}\right) &= \langle k(\cdot,\boldsymbol{z}),\, \frac{\partial k(\cdot,\boldsymbol{z}')}{\partial z'_j}\rangle_{\mathcal{H}_k} = \frac{\partial k(\boldsymbol{z},\boldsymbol{z}')}{\partial z'_j}, \label{eq:cross_cov}\\
    \operatorname{Cov}\!\left(\frac{\partial f_d(\boldsymbol{z})}{\partial z_i},\, \frac{\partial f_d(\boldsymbol{z}')}{\partial z'_j}\right) &= \left\langle \frac{\partial k(\cdot,\boldsymbol{z})}{\partial z_i},\, \frac{\partial k(\cdot,\boldsymbol{z}')}{\partial z'_j}\right\rangle_{\mathcal{H}_k} = \frac{\partial^2 k(\boldsymbol{z},\boldsymbol{z}')}{\partial z_i\,\partial z'_j}. \label{eq:deriv_cov}
\end{align}
which thus also inherit RKHS-associated positive semi-definiteness and constitute  valid covariance kernels.

Therefore, the concatenated vector  $(\boldsymbol{u}_d, \boldsymbol{v}_d^{\theta})$ of the GP and its derivative
process is still jointly distributed as a GP with covariance $\mathsf{K}_{\zeta\zeta}^{\theta}$ 
\begin{equation}
    \begin{pmatrix} \boldsymbol{u}_d \\ \boldsymbol{v}_d^{\theta} \end{pmatrix} \sim \mathcal{N}\!\left(\boldsymbol{0},\; \mathsf{K}_{\zeta\zeta}^{\theta}\right)
\end{equation}
where $\mathsf{K}_{\zeta\zeta}^{\theta}$ is given by 
\begin{equation}
    \mathsf{K}_{\zeta\zeta}^{\theta} =
    \begin{bmatrix}
        K_{\zeta\zeta} & K_{\zeta\mathcal{S}} & \nabla_2 K_{\zeta\mathcal{S}} \\
        K_{\mathcal{S}\zeta} & K_{\mathcal{S}\mathcal{S}} & \nabla_2 K_{\mathcal{S}\mathcal{S}} \\
        \nabla_1 K_{\mathcal{S}\zeta} & \nabla_1 K_{\mathcal{S}\mathcal{S}} & \nabla_1\nabla_2 K_{\mathcal{S}\mathcal{S}}
    \end{bmatrix}
\end{equation}
with $\nabla_i$ denoting differentiation with respect to the $i$-th argument. Conditioning $\boldsymbol{u}_d$ on $\boldsymbol{v}_d^{\theta}$ then yields the prior in Section~\ref{sec: interpretable GPs}.

\section{Algorithmic Details}
\begin{algorithm}[h]
\caption{Learning RP-GPdyn conditioned on fixed points and their associated Jacobians}
\label{alg:rpgpdyn}
\begin{algorithmic}[1]
\Require Dataset $\mathbb{X}^{(N)}=\{\mathcal{X}^{(n)}\}_{n=1}^N$, with $\mathcal{X}^{(n)}=\{\v{x}_{t}^{(n)}\in\mathbb{R}^{P}\}_{t=1}^{T}$; kernel hyperparameters;
\State Initialize \\
\hspace{\algorithmicindent} recognition parameters $\phi$,\\
\hspace{\algorithmicindent} variational posterior parameters $\psi$,\\
\hspace{\algorithmicindent} inducing locations $\m{\zeta}=\{\v{\zeta}_m\}_{m=1}^{M}$,\\
\hspace{\algorithmicindent} fixed point locations $\mathcal{S}=\{\v{s}_{\ell}\}_{\ell=1}^{L}$,\\
\hspace{\algorithmicindent} Jacobians $\{\m{J}^{(\ell)}\}_{\ell=1}^{L}$\\
\hspace{\algorithmicindent} transition variance $\sigma^2$,
\For{$i=1,\ldots,K$}
    \State Sample a mini-batch $\mathcal{B}\subset\{1,\ldots,N\}$

    \ForAll{$n\in\mathcal{B}$}
        \State Compute recognition factors $g_{\phi}(\v{z}_t\mid \v{x}_{t}^{(n)})$, for $t=1,\ldots,T$
        \State Compute amortised variational posterior parameters with a bidirectional recurrent network
        \Statex \hspace{4.5em} $\{\m{A}_{t}^{(n)}, \v{b}_{t}^{(n)}, \m{Q}_{t}^{(n)}\}_{t=1}^T = \operatorname{BiRNN}_{\psi}(\v{x}_{1:T}^{(n)})$ 
        \Statex \hspace{4.5em} $q_{\psi}^{n}(\v{z}_1)=\mathcal{N}(\v{\mu}_{0}^{(n)},\m{Q}_{0}^{(n)})$
        \Statex \hspace{4.5em} $q_{\psi}^{n}(\v{z}_t\mid \v{z}_{t-1})
        =
        \mathcal{N}\!\left(
        \m{A}_{t}^{(n)}\v{z}_{t-1}+\v{b}_{t}^{(n)},
        \m{Q}_{t}^{(n)}
        \right)$, for $t=2,\ldots,T$
    \EndFor
    \State Form the mini-batch interior variational bound $\tilde{\mathcal{F}}$ based on Eq.~\ref{eq: final aux free energy}
    \State Update variational GP quantities 
    $(\v{m}_d,\m{\Gamma}_d, \{\m{J}^{(\ell)}\}_{\ell=1}^{L})
    \leftarrow
    \operatorname{argmax} \tilde{\mathcal{F}}
    $ based on Eq.~\ref{eq:closed_form_gamma}-\ref{eq:closed_form_M}

    \State Take a stochastic gradient step on the remaining parameters
    \Statex \hspace{\algorithmicindent}
    $(\phi,\psi,\m{\zeta},\mathcal{S},\sigma^2)_i
    \leftarrow
    \mathrm{Adam}\!\left(
    (\phi,\psi,\m{\zeta},\mathcal{S},\sigma^2)_{i-1}, \;
    \nabla\widehat{\mathcal{F}}_{\mathrm{int}}
    \right)$
\EndFor
\end{algorithmic}
\end{algorithm}

\section{Additional Synthetic Data Experimental Details}

\label{app:synthetic_history_effects}

For Figure \ref{fig: main vdp spikes}, we generate two-dimensional ground-truth latent trajectories from a Van~der~Pol differential equation with deterministic drift
\begin{align}
\dot z_1^* &= \tau a\left(z_1^*-\frac{1}{3}\left(z_1^*\right)^3-z_2^*\right), \\
\dot z_2^* &= \tau\frac{z_1^*}{a},
\end{align}
where \(a=2\) and \(\tau=20\). We simulate trajectories at $\Delta t_{\mathrm{sim}} =1$ ms resolution using Euler discretisation, generating 50 training trials and 30 held-out test trials.

\begin{figure}
    \centering
    \includegraphics[width=\linewidth]{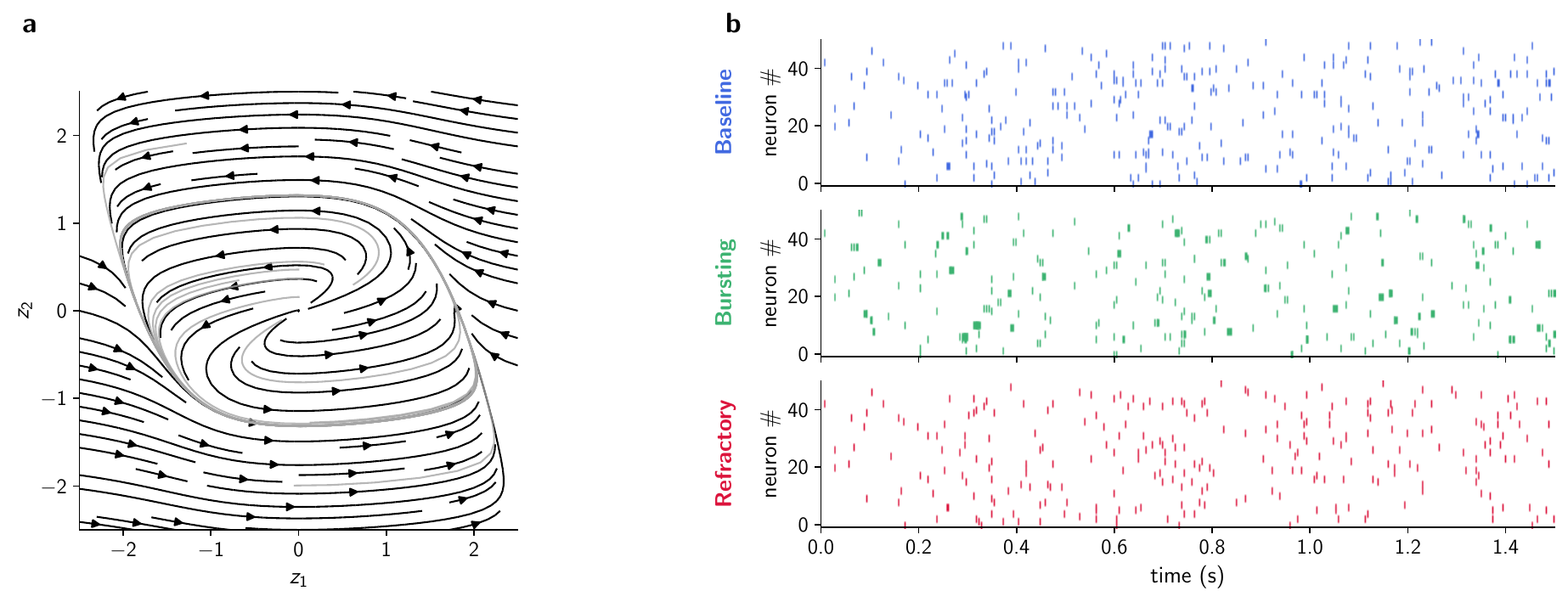}
    \caption{\textbf{(a)} Ground truth Van der Pol flow field and trajectories. \textbf{(b)} Simulated activity for baseline, bursting, and refractory conditions. }
    \label{fig: history description}
\end{figure}

Each ground-truth latent state \(\boldsymbol{z}^{*(n)}_k\) is linearly mapped to the drive of 50 neurons
\begin{equation}
u^{(n)}_{k,j}
=
\boldsymbol{c}_j^\top\boldsymbol{z}^{*(n)}_k+d_j.
\end{equation}
where \[
\boldsymbol z_k^{*(n)}=\boldsymbol z^{*(n)}(t_k),
\qquad
t_k=k\Delta t_{\mathrm{sim}}.
\] The readout is sampled once and shared across trials and data splits. In the history-free setting, the spike count of neuron \(j\) in each 1 ms interval is sampled as
\begin{equation}
y^{(n)}_{k,j}
\sim
\mathrm{Poisson}\!\left[
\Delta t_{\mathrm{sim}}\,
\mathrm{softplus}\!\left(u^{(n)}_{k,j}\right)
\right],
\end{equation}
where \(\Delta t_{\mathrm{sim}}=1\) ms. To introduce temporal effects, we add a causal self-history term before the softplus nonlinearity:
\begin{equation}
y^{(n)}_{k,j}
\sim
\mathrm{Poisson}\!\left[
\Delta t_{\mathrm{sim}}\,
\mathrm{softplus}\!\left(
u^{(n)}_{k,j}
+
\sum_{\ell=1}^{L_h}
\kappa_\ell y^{(n)}_{k-\ell,j}
\right)
\right].
\end{equation}
Here, \(L_h\) is the number of history lags and \(\kappa_\ell\) is the coefficient at lag \(\ell\). Refractory filters are negative and exponentially decaying, causing a spike to suppress subsequent firing transiently. Bursting filters contain a positive short-lag component followed by a delayed negative component, promoting closely spaced spikes before suppressing later activity. Finally, we sum the 1 ms spike counts within non-overlapping 5 ms bins to obtain the observations (1.5 seconds total) supplied to the models.
Figure \ref{fig: history description} shows the resulting data and history effects.


For the present experiments, RP-GPdyn uses a two-dimensional latent state and a sparse RBF-GP prior with 30 inducing points and one fixed point. The inducing locations are initialised uniformly in \([-2,2]^2\) and then optimised, together with the remaining model parameters, by minimizing the auxiliary free energy. The RBF length scale is initialised to \(0.1\) and learned. The inducing-variable posterior is updated in closed form at each optimisation step. The fixed-point location is initialised separately in \([-0.5,0.5]^2\) and is likewise trainable. The recognition factors use an inverse-softplus transform followed by a learned linear map. The temporal variational posterior is parameterised by a bidirectional GRU with 64 hidden units per direction. We optimise the variational free energy with Adam using minibatches of size 32 for 10,000 updates, with global gradient clipping at norm \(1.0\) and a learning rate that decays linearly from \(10^{-3}\) to \(10^{-4}\).

SING was implemented using the official repository associated with \cite{hu2026sing} (https://github.com/lindermanlab/sing). We used a softplus-Poisson count likelihood, and an RBF-GP with 81 inducing points on a \(9\times9\) grid spanning \([-2.25,2.25]^2\). The length scale was initialised to \(1.0\) and learned. A softplus-Poisson LDS initialised the observation map, similar to \citet{hu2024modeling}. We ran 50 full-batch variational-EM iterations, each with 30 E-step and 50 M-step updates, using an M-step learning rate of \(10^{-4}\). 

Before evaluation, each model's latent coordinates were aligned to the ground-truth space using a least-squares linear map fitted on the training trials and held fixed for the test trials. For each test trial \(n\) and latent dimension \(d\), we computed
\[
R^2_{n,d}
=
1-
\frac{\sum_{k=k_0}^{K}\left(\hat z_{k,d}^{(n)}-z_{k,d}^{*(n)}\right)^2}
{\sum_{k=k_0}^{K}\left(z_{k,d}^{*(n)}-\bar z_d^{*(n)}\right)^2}
\]
where \(\bar z_d^{*(n)}\) is the ground-truth mean over the same time points, and \(R^2\) was averaged across latent dimensions. For inference, \(\hat z=\langle z\rangle_q\). For prediction, the rollout was initialised at the ground-truth state following a \(0.5\,\mathrm{s}\) prefix and recursively propagated using the posterior mean dynamics; \(R^2\) was then computed over the remaining future time points.


Table \ref{tab:compute} compares runtime for RP-GPdyn and SING. The computational scaling with latent dimensionality and the number of inducing points is inherited from the GP dynamical predictor and is therefore not expected to differ substantially between methods. We define an iteration of RP-GPdyn as a single gradient update in the parameters.  This is performed using reverse-mode automatic differentiation (\texttt{jax.value and grad}), and thus includes the compute required for inference by running the bidirectional RNN latent estimator. SING has a separate inference (E-) step, and training used different numbers of E- and M-step iterations. Here we have averaged the runtime per single (E- or M-) step in the proportion that they were used during learning. Although models and hardware requirements are not directly comparable, the improved performance of RP-GPdyn came at lower overall compute cost and runtime.

\begin{table}[t]
\centering
\caption{Training configuration and runtime per iteration.}
\label{tab:compute}
\small
\begin{tabular}{lcccc}
\toprule
Model & Inducing & Optimisation schedule & GPU & Runtime / iteration (s) \\
\midrule
RP-GPdyn
& 30
& 10k grad. updates, batch 32
& T1000 (Turing)
& $0.4285 \pm 0.0003$ \\

SING
& 81
& 50 vEM; 30 E + 50 M, full batch
& RTX 4500 (Ada)
& $2.453 \pm 0.059$ \\
\bottomrule
\end{tabular}
\end{table}


\section{Additional Decision Making Experimental Details}\label{ap:decision-making}

Neural activity was simulated by the spiking EI network model of \citet{wang2002probabilistic} with 1600 excitatory neurons (1120 non-selective, 240 left-selective, 240 right-selective) and 400 inhibitory neurons. The dataset consisted of $N=2193$ trials, with 11 coherence levels ($\sim200$ trials per level). Each trial is $2$\,s long, with stimulus onset at $t=0.5$\,s and stimulus offset at $t=1.5$\,s (giving $0.5$\,s of pre- and post-stimulus periods, the first of which was discarded from analysis). The decision outcome of each trial was determined by the sign of the difference in average firing rate of the two selective populations in the post-stimulus period, thresholded at 5\,Hz (109 trials with difference magnitude smaller than 5\,Hz were marked ``indecisive'').

Observations to RP-GPdyn consisted of population vectors of spike counts of the excitatory population from stimulus onset, $t\in\left[0.5,2\right]$, binned to 10\,ms, resulting in 150 timesteps per trial. The recognition factors are parameterised by an MLP with one hidden layer of 32 neurons and ReLU activation function that maps the neural observations to the mean of a two-dimensional Gaussian, and an input-independent diagonal covariance matrix. 
To model the ``shared'' mode of the dynamics among different trials, we used the time-pooled reference measure $p_0(x_t)$.  

\subsection{Evaluation}

\paragraph{Qualitative evaluation} We computed $q(f)$ under the trained model, conditioned on the training split (see below). We then plotted the inferred flow-field as the mean one-step displacement function  $\mathbb{E}_{q}\left[f(z)\right]-z$, as well as the associated uncertainty through the variance of $q(f)$, summed over the two latent dimensions.
Next, we plotted the posterior means of inferred latent trajectories in the same space, coloured by either the trial coherence level or the trial choice (for ambiguous 0-coherence trials).

\paragraph{Neural activity prediction} For training a readout probe, we randomly split the dataset into training ($n=1096$) and validation ($n=1097$) trials. We trained a ridge regression from RBF features of the two-dimensional posterior mean to predict the instantaneous left-, right- and non-selective population firing rates, centred on their training mean. The readout is instantaneous, trained by pooling over all time bins in the train split. The ridge coefficient was set to $\lambda=100$; held-out $R^2$ within the train split was insensitive to $\lambda\in\left[1,100\right]$, with larger values used to suppress oscillations of the map away from data.
We visualized this probe by plotting, across the latent space, its prediction of $\text{R}-\text{L}$.

The trained readout was then used to evaluate how well neural activity can be forecast. For that, we inferred posteriors over latent states for validation trials, given only a prefix of $K=20,30,\ldots,100$ observations (0.2\,s, 0.3\,s, \ldots, 1\,s). States for the remaining $150-K$ time steps were then computed by rollouts: we sampled 8 particles from $q(z_{t_\text{prefix}})$, and then iteratively sampled $z_{t+1} \sim \mathcal{N}(\mu_q(z_t), \text{diag}(\sigma^2_q(z_t)) + \sigma^2I)$, where $\mu_q$ and $\sigma^2_q$ are the predictive mean and variance of the GP transition, and $\sigma$ is the (learned) transition-noise standard deviation (see Eq.~\eqref{eq: f prior}). Neural predictions were then obtained by applying the trained probe at each particle along the rollout, and averaging the predictions at each time-step over the 8 particles. We compare to two controls:
\begin{enumerate}
    \item \textbf{Frozen:} the initial particles sampled from $q(z_{t_\text{prefix}})$ are used throughout the rollout, without updating them.
    \item \textbf{Full posterior:} rather than rollouts, we compute the full posterior over the trial, conditioning on observations beyond the prefix. Neural predictions are then computed directly from the posterior means.
\end{enumerate}

For each condition (rollout, frozen, full) and each prefix, we computed the per-population (L/R) $R^2$ relative to the training-split PSTH:
$$R^2_p = 1 - \frac{\sum_{n \in \mathcal{V}} \sum_{t=K}^{T-1} \left(y_{ntp} - \hat{y}_{ntp}\right)^2}{\sum_{n \in \mathcal{V}} \sum_{t=K}^{T-1} \left(y_{ntp} -
  \bar{y}_{tp}\right)^2}, \qquad p \in \lbrace \mathrm{L}, \mathrm{R}\rbrace  $$
where $y_{ntp}$ is the observed firing rate of population $p$ in validation trial $n$ at time-step $t$, $\hat{y}_{ntp}$ is the predicted firing rate at the same indices, and $\bar{y}_{tp}$ is the \textit{training} PSTH pooled over all coherence levels. We then averaged $R^2_R$ and $R^2_L$ to report a single performance measure. 
Note that each prefix is scored on a different window (the corresponding post-prefix time steps).

\paragraph{Choice prediction} We used the same rollouts to also predict trial choice, as
$$\text{predicted choice} =\text{sign}\left(\frac{1}{20}\sum_{t\ge130}\left(\hat{y}_{ntR}-\hat{y}_{ntL}\right)\right),$$ 
the sign of the predicted difference $\text{R}-\text{L}$, averaged over the last 20 time-steps. 
We computed the prediction accuracy by comparing to the true trial outcomes computed from the neural activity (with ``undecided'' trials excluded for this analysis). As a baseline, we computed the fraction of (decided) $\lvert c \rvert > 0$ trials whose outcome is congruent with the signed coherence (0.87). Prediction accuracy above this baseline for non-zero coherence trials indicates that RP-GPdyn can predict actual trial outcomes also in ``error'' trials (network choice incongruent with the stimulus coherence). For zero coherence trials, choice prediction accuracy was considerably above naive $0.5$ chance.

\section{Additional \texttt{MC\_Maze} Experimental Details}\label{ap: mcmaze}

\subsection{Dataset and evaluation settings}

We used the MC\_Maze dataset distributed through the Neural Latents Benchmark
(NLB), comprising simultaneous recordings from 182 neurons in primary motor
and dorsal premotor cortex, together with hand position, hand velocity and
trial metadata. Conditions were defined by the pair
\texttt{(trial\_type, trial\_version)}, yielding 108 reach conditions.
We distinguished two evaluation settings. For neural reconstruction
(Figure~5g), we used the official NLB split: 1,721 training trials, 574
validation trials and a separate set of 574 test trials. Of the 182 neurons,
137 were designated held-in and 45 held-out. For generalisation across
movements (Figure~5h,i), we instead split the 2,295 public training and
validation trials by reach condition. With random seed 10, 98 conditions
(2,082 trials) formed the training set and the remaining 10 conditions
(213 trials) formed the outer evaluation set. This condition split is
distinct from the official NLB test set. 

\subsection{Neural preprocessing and alignment}

Spike counts were binned in non-overlapping 20\,ms intervals and aligned to
movement onset. Neural observations were smoothed separately within each
trial using a normalised Gaussian kernel with standard deviation 70\,ms
and support approximately $\pm3$ standard deviations. For the public
portable export, smoothing preceded extraction of the aligned window,
retaining the available within-trial context. Smoothing did not mix trials.
For official test inputs, only the supplied 700\,ms neural window was
available, and smoothing used zero padding outside this window. Smoothing
was noncausal.

The condition-generalisation models were trained on
$[-200,700)$\,ms relative to movement onset (45 bins). The NLB models were trained on the portable export's $[-240,460)$\,ms window (35 bins).
Official evaluation used the NLB window nominally specified as
$[-250,450)$\,ms, also containing 35 bins at 20\,ms resolution. Because
movement onset need not lie on a bin boundary, the official tensor clock
and the portable movement-onset clock can differ by one bin. We preserved
the official trial order, channel order and per-trial bin alignment when
constructing evaluation targets.

Preprocessing depended on the saved model configuration. Let $s_{ntp}$
denote the smoothed count of neuron $p$ in trial $n$ at time $t$.

All NLB models in
Figure~5g and the eight-dimensional models in Figure~5h,i used a single
mean per neuron, pooled over trials and time:
\begin{equation}
 x_{ntp}=s_{ntp}-\mu_p,\qquad
 \mu_p=\frac{1}{NT}\sum_{n,t}s_{ntp}.
\end{equation}
Following \cite{churchland2012neural}, the
two-dimensional flow examples in Figure~5c--f in addition used soft normalisation
and subtraction of the across-trial mean at each time bin:
\begin{align}
 r_p &= \max_{n,t}s_{ntp}-\min_{n,t}s_{ntp}, \\
 x_{ntp} &= \frac{s_{ntp}}{r_p+5}
 -\frac{1}{N}\sum_{n'=1}^{N}\frac{s_{n'tp}}{r_p+5}.
\end{align}

\subsection{Latent models and optimisation}

All Maze models used a sparse squared-exponential GP transition and an
amortised Gaussian Markov posterior. The posterior network was a
bidirectional GRU with 64 hidden units in each direction, followed by
projections to the Gaussian-chain parameters. Inference therefore used
the neural observations throughout the input window. Linear recognition
used an affine projection, whereas nonlinear recognition used fully
connected ReLU layers followed by an affine projection. Recognition
factors had time-invariant diagonal precision. 

Initial GP length scale, kernel amplitude and latent transition noise
variance were $0.1$, $1.0$ and $1.0$, respectively. Inducing locations
were initialised uniformly in $[-1,1]^D$. No Maze model listed here used
fixed-point or Jacobian constraints. Optimisation used Adam with constant
learning rate $10^{-3}$, batches of 64 trials, and global gradient-norm limit
1. 
The kernel amplitude was held fixed. Length scale and transition noise were optimised. 

Two empirical reference measures were considered. The time-specific
measure pools observations across sequences at the same time bin,
whereas the time-pooled measure pools over both sequences and time.
The NLB sweeps used all time bins for time pooling. 


\subsection{Flow fields and condition-averaged latent trajectories}

For the two-dimensional examples, the displayed vector field was the
posterior-mean one-step displacement $\mathbb{E}[f(z)]-z$ of the learned
GP transition. Background shading represented the summed predictive
variance over latent coordinates. To obtain the displayed PSTH-driven
latent trajectories, we first averaged preprocessed neural observations
within each reach condition and then applied the amortised posterior to
each condition-averaged input sequence. Colours
were determined by mean measured hand-velocity direction over
$[0,300)$\,ms after movement onset. Inferring a trajectory from a
condition-averaged input is distinct from averaging posterior trajectories
inferred from individual trials, which is the procedure underlying the
neural prediction scores below.

\subsection{Neural reconstruction}

The NLB sweep used $D\in\{2,4,6,8\}$, $M=60$ inducing locations and
30,000 latent-model updates. During training, the recognition factor
received all 182 neurons, while the amortised posterior received only
the 137 held-in neurons. At test time, the posterior likewise received
only held-in neural activity. Behaviour was not used to train these
models. 


For each frozen model, an external instantaneous neural readout mapped
posterior means to 182-channel neural predictions. The readout consisted
of a 64-unit ReLU hidden layer and a linear output layer. Inputs and
smoothed single-trial neural targets were standardised using fitting-set
means and standard deviations, with a standard-deviation floor of
$10^{-6}$. Training minimised squared error in standardised target
coordinates using Adam with learning rate $3\times10^{-3}$, batch size
512 and global gradient-norm limit 1. Predictions were transformed back
to neural count units, restoring the target mean once. 

The stopping step was selected using the 1,721
training and 574 validation trials, with initialisation seed 30. Validation
single-trial neural $R^2$ was checked every 100 updates, for at most
5,000 updates. A fresh readout was then fitted
on all 2,295 public trials for the selected number of updates using seed
130. 
The same posterior-trained readout, including its scaling statistics, was
reused unchanged for rollout evaluation. Official
test targets were excluded from readout optimisation and stopping rules.

For scoring, single-trial neural predictions were averaged using the
official condition membership and compared with the official NLB test
PSTHs. If $y_{ctp}$ and $\hat y_{ctp}$ denote the reference and predicted
PSTHs, respectively, we computed
\begin{equation}
 R^2_p=1-\frac{\sum_{c,t}(y_{ctp}-\hat y_{ctp})^2}
 {\sum_{c,t}(y_{ctp}-\bar y_p)^2},\qquad
 \bar y_p=\frac{1}{CT}\sum_{c,t}y_{ctp},
 \quad R^2_{\mathrm{PSTH}}=\frac{1}{182}\sum_{p=1}^{182}R^2_p.
\end{equation}
Thus, Figure~5g reports an unweighted mean across all 182 neurons,
including both held-in and held-out neurons, over 108 conditions and
35 time bins. Negative neuron scores were retained. This calculation
agreed with the official NLB PSTH evaluator. 

\subsection{GP rollout reconstruction}

Rollouts used the frozen GP transition parameters. The variational GP
state used for prediction was recomputed from the original 1,721
training trials, without refitting the transition on validation or test
trials. Let $t_0$ denote movement onset on the relevant trial's bin grid
and $m_{nt}$ the full-window posterior mean. We constructed
\begin{equation}
 \tilde z_{nt}=m_{nt}\quad(t\leq t_0),\qquad
 \tilde z_{n,t+1}=\mathbb{E}[f(\tilde z_{nt})]\quad(t\geq t_0).
\end{equation}
The post-onset path therefore propagated the GP transition mean without
stochastic transition samples or subsequent observation updates. The
pre-onset posterior prefix was retained and the full 35-bin window was
scored. Rollout scores reused the corresponding posterior-trained neural readout,
including its input and target scaling statistics. 

\subsection{Behaviour-guided latent training}

The two-arm models used separate recognition factors for neural activity
and two-dimensional hand velocity, with observation dimensions 182 and 2.
Their product entered the training objective under conditional
independence of the two modalities given the latent state. The velocity
stream was shifted forward by five 20\,ms bins relative to the neural
stream, pairing neural activity at time $t$ with velocity at
$t+100$\,ms. The velocity input was not Gaussian-smoothed or subjected
to the neural soft-normalisation procedure. Behavioural channels were
excluded from the amortised posterior at both training and evaluation:
the auxiliary velocity factor shaped latent-model learning, but
posterior trajectories could be inferred from neural activity alone.
Figure~5f uses the two-dimensional time-specific model, whereas
Figure~5i compares eight-dimensional time-pooled models.

\subsection{Behavioural readouts and held-out-condition scoring}

For Figure~5h,i, posterior means were recomputed from neural input
windows $[-100,500)$\,ms relative to movement onset (30 bins), using
each checkpoint's saved preprocessing. These condition-split models
received all 182 neural channels. We decoded both position and velocity with a
five-bin offset. The latent inputs used for decoding spanned
$0,20,\ldots,380$\,ms, and the corresponding targets spanned
$100,120,\ldots,480$\,ms. The lag was applied once to the kinematic
targets; it was not an additional shift of already lag-aligned targets.

Each latent model and each target had separately fitted readouts. The
instantaneous MLP used a 64-unit ReLU layer and a two-coordinate linear
output. The causal RNN used a single 64-unit recurrent layer
and a linear output, with its hidden state initialised to zero at the
first post-onset latent input. Both readouts used standardised inputs
and targets, masked squared-error loss, Adam with learning rate
$3\times10^{-3}$ and global gradient-norm limit 1. The MLP used batches
of 256 samples and at most 250 epochs; the RNN used batches of 128
trials and at most 120 epochs. Checkpoints were evaluated at the first
epoch and every five epochs thereafter, retaining the best development
MSE, with patience 15 checks for the MLP and 10 for the RNN. Scaling
statistics were estimated using only the inner fitting subset, and
the selected readouts were evaluated on the untouched outer conditions.
The RNN readout was causal in its latent inputs, but the full decoding
pipeline remained offline because the latent posterior was bidirectional
and the neural preprocessing included noncausal smoothing.

The behavioural score pooled all valid trial--time samples and both
coordinates. For $j\in\{x,y\}$ and valid sample index $a$,
\begin{equation}
 R^2_{\mathrm{beh}}=1-
 \frac{\sum_{a,j}(y_{aj}-\hat y_{aj})^2}
 {\sum_{a,j}(y_{aj}-\bar y_j)^2},\qquad
 \bar y_j=\frac{1}{A}\sum_a y_{aj}.
\end{equation}
This is a variance-weighted aggregation of coordinate-wise $R^2$,
rather than an unweighted mean of the two coordinate scores.
Figure~5h reports single-trial direct velocity and direct position
decoding. These scores use the condition split and MLP/RNN readouts
specified above, and are potentially stricter than the official NLB ridge velocity score since there is no condition splitting in NLB.

For Figure~5i, the MLP-predicted velocity was integrated at $\Delta t=0.020$\,s:
\begin{equation}
 \hat p_0=p_0,\qquad
 \hat p_k=p_0+\Delta t\sum_{\ell=0}^{k-1}\hat v_\ell.
\end{equation}
Here $p_0$ was the measured position at the first target time
(100\,ms after movement onset); no later position measurements reset
the integration. Both reference and reconstructed paths were translated
by this same initial position for display. The examples show the two
selected outer-evaluation reach conditions with target coordinates
$(123,71)$ and $(129,-29)$ in the export's position units, corresponding
to \texttt{(trial\_type, trial\_version)} pairs $(19,2)$ and $(39,1)$.
The displayed velocity $R^2$ values, 0.702 for neural-only training and
0.814 for two-arm training, were computed over the entire outer
evaluation set, not only the displayed examples. They score velocity,
not the integrated hand paths.

\section{Additional ALM Experimental Details}\label{ap: ALM}

For the ALM experiments, we analysed delay-period spike-count activity from four fixed-delay sessions ( \texttt{HI124\_020817}, \texttt{HI127\_031517}, \texttt{HI129\_040517}, \texttt{HI129\_040717}) obtained at \url{https://doi.org/10.25378/janelia.7489253}. Counts were aggregated into 70 ms bins. For each session and model, trials were split into 80\% training and 20\% held-out test trials. All recorded neurons were retained in both splits. The example flow fields and trajectories use \texttt{HI129\_040517}, whereas the choice-decoding analysis includes all four sessions.

RP-GPdyn uses a time-pooled base measure and a two-dimensional latent state and a sparse RBF-GP prior with 30 inducing points. The inducing locations are initialised uniformly in \([-2,2]^2\). Additionally, three fixed points are initialised independently in \([-0.5,0.5]^2\). The RBF length scale is initialised to \(0.1\). The temporal variational posterior is parameterised by a bidirectional GRU with 32 hidden units per direction. We minimise the auxiliary free energy using Adam for 10,000 updates, with minibatches of 16 trials, global gradient clipping at norm \(5.0\), and a learning rate of \(10^{-3}\). 



For SING we used a two-dimensional latent state, a softplus-Poisson count likelihood, and an RBF-GP with 144 inducing points on a \(12\times12\) grid spanning \([-4,4]^2\). The length scale was initialised to \(0.5\) and learned. A softplus-Poisson LDS  initialised the observation map. We ran 50 full-batch variational-EM iterations, each with 30 E-step and 50 M-step updates, using an M-step learning rate of \(10^{-4}\). 

To quantify choice accuracy, we defined the prefix as the first $0.4$s of the observed spike-count sequences. For each held-out trial, latent inference was rerun using only these counts. Its final posterior mean (i.e., inference) initialised a recursive mean-dynamics rollout (i.e., prediction) to the trial end. A ridge decoder mapped the rollout states to predicted neural activity. A logistic classifier then predicted left versus right choice from the predicted activity averaged over the final four bins. Accuracy was the fraction of held-out trials classified correctly.

\section*{NeurIPS Paper Checklist}

\begin{enumerate}

\item {\bf Claims}
    \item[] Question: Do the main claims made in the abstract and introduction accurately reflect the paper's contributions and scope?
    \item[] Answer: \answerYes{} 
    \item[] Justification:
The abstract and introduction describe the paper's primary contribution as a recognition-parameterized Gaussian-process dynamical model for recovering interpretable nonlinear latent dynamics from neural recordings without requiring explicit generative observation modeling. These claims are supported through experiments on synthetic dynamical systems and neural datasets, together with analyses of latent trajectories, fixed points, and learned phase portraits. The discussion section also explicitly states the scope and limitations of the current framework.


\item {\bf Limitations}
    \item[] Question: Does the paper discuss the limitations of the work performed by the authors?
    \item[] Answer: \answerYes{} 
    \item[] Justification: The paper explicitly discusses limitations of the proposed approach, including assumptions of autonomous latent dynamics, the absence of explicit exogenous inputs, computational considerations associated with GP-based inference, and potential challenges related to scalability and identifiability. 


\item {\bf Theory assumptions and proofs}
    \item[] Question: For each theoretical result, does the paper provide the full set of assumptions and a complete (and correct) proof?
    \item[] Answer: \answerYes{} 
    \item[] Justification: The paper provides complete derivations for the proposed probabilistic modeling framework and variational inference procedure, including the assumptions underlying the Gaussian-process dynamical model, recognition parameterization, and approximate inference formulation. Mathematical details and derivations are included in the main text and supplementary material.

    \item {\bf Experimental result reproducibility}
    \item[] Question: Does the paper fully disclose all the information needed to reproduce the main experimental results of the paper to the extent that it affects the main claims and/or conclusions of the paper (regardless of whether the code and data are provided or not)?
    \item[] Answer: \answerYes{} 
    \item[] Justification: Experimental settings, datasets, baselines, and evaluation procedures are described in the main text and supplementary material. 

\item {\bf Open access to data and code}
    \item[] Question: Does the paper provide open access to the data and code, with sufficient instructions to faithfully reproduce the main experimental results, as described in supplemental material?
    \item[] Answer: \answerYes{} 
    \item[] Justification: The implementation is written in JAX and the codebase will be released upon publication to support reproducibility. The supplementary material additionally documents model configurations, optimization procedures, and experimental settings required to reproduce the reported results.


\item {\bf Experimental setting/details}
    \item[] Question: Does the paper specify all the training and test details (e.g., data splits, hyperparameters, how they were chosen, type of optimizer) necessary to understand the results?
    \item[] Answer: \answerYes{} 
    \item[] Justification: The supplementary material includes implementation details, hyperparameters, preprocessing, and training procedures.

\item {\bf Experiment statistical significance}
    \item[] Question: Does the paper report error bars suitably and correctly defined or other appropriate information about the statistical significance of the experiments?
    \item[] Answer: \answerYes{} 
    \item[] Justification: Quantitative results include uncertainty estimates.

\item {\bf Experiments compute resources}
    \item[] Question: For each experiment, does the paper provide sufficient information on the computer resources (type of compute workers, memory, time of execution) needed to reproduce the experiments?
    \item[] Answer: \answerYes{} 
    \item[] Justification: Reported in the supplementary material. 
    
\item {\bf Code of ethics}
    \item[] Question: Does the research conducted in the paper conform, in every respect, with the NeurIPS Code of Ethics \url{https://neurips.cc/public/EthicsGuidelines}?
    \item[] Answer: \answerYes{}
    \item[] Justification: The work uses synthetic and previously published neural datasets and does not introduce apparent violations of the NeurIPS Code of Ethics. 

\item {\bf Broader impacts}
    \item[] Question: Does the paper discuss both potential positive societal impacts and negative societal impacts of the work performed?
    \item[] Answer: \answerYes{} 
    \item[] Justification: The paper discusses the scientific utility and limitations of interpretable neural dynamical models.
    
\item {\bf Safeguards}
    \item[] Question: Does the paper describe safeguards that have been put in place for responsible release of data or models that have a high risk for misuse (e.g., pre-trained language models, image generators, or scraped datasets)?
    \item[] Answer: \answerNA{} 
    \item[] Justification: The paper does not release a deployed system or model requiring misuse safeguards.

\item {\bf Licenses for existing assets}
    \item[] Question: Are the creators or original owners of assets (e.g., code, data, models), used in the paper, properly credited and are the license and terms of use explicitly mentioned and properly respected?
    \item[] Answer: \answerYes{} 
    \item[] Justification: Existing datasets and software dependencies are cited, with licenses provided where available.

\item {\bf New assets}
    \item[] Question: Are new assets introduced in the paper well documented and is the documentation provided alongside the assets?
    \item[] Answer: \answerYes{}
    \item[] Justification:  The released code will be documented with usage instructions and implementation details.

\item {\bf Crowdsourcing and research with human subjects}
    \item[] Question: For crowdsourcing experiments and research with human subjects, does the paper include the full text of instructions given to participants and screenshots, if applicable, as well as details about compensation (if any)? 
    \item[] Answer: \answerNA{}
    \item[] Justification: The work does not involve crowdsourcing or new human-subject experiments.

\item {\bf Institutional review board (IRB) approvals or equivalent for research with human subjects}
    \item[] Question: Does the paper describe potential risks incurred by study participants, whether such risks were disclosed to the subjects, and whether Institutional Review Board (IRB) approvals (or an equivalent approval/review based on the requirements of your country or institution) were obtained?
    \item[] Answer: \answerNA{} 
    \item[] Justification:  The work does not involve new human-subject research requiring IRB approval.

\item {\bf Declaration of LLM usage}
    \item[] Question: Does the paper describe the usage of LLMs if it is an important, original, or non-standard component of the core methods in this research? Note that if the LLM is used only for writing, editing, or formatting purposes and does \emph{not} impact the core methodology, scientific rigor, or originality of the research, declaration is not required.
    \item[] Answer: \answerNA{}
    \item[] Justification:  LLMs are not used as an important, original, or non-standard component of the core methodology.

\end{enumerate}
\end{document}